\documentclass[11pt]{article}
\usepackage{amsfonts}
\usepackage{tabularx}
\usepackage{xcolor}
\usepackage{array}
\usepackage{wrapfig}
\usepackage{booktabs}
\usepackage{enumitem}
\usepackage{framed}
\usepackage{fancyvrb}
\usepackage{tcolorbox}
\tcbuselibrary{breakable}
\usepackage[normalem]{ulem}
\usepackage{subcaption}

\usepackage[final]{acl}
\definecolor{darkgreen}{RGB}{0,128,0}

\usepackage{graphicx}
\usepackage{amsmath}
\usepackage{xcolor}

\definecolor{mypink}{HTML}{B85450}
\definecolor{mypurple}{HTML}{9673A6}
\definecolor{myblue}{HTML}{6C8EBF}
\definecolor{mygreen}{HTML}{82B366}
\definecolor{myorange}{HTML}{D79B00}
\definecolor{grey}{HTML}{E6E6E6}
\definecolor{darkgreen}{RGB}{0,128,0}
\definecolor{purple}{RGB}{128,0,128}
\definecolor{graytext}{RGB}{100,100,100}
\definecolor{blue}{RGB}{0,0,200}
\usepackage{times}
\usepackage{latexsym}
\usepackage{multirow}

\usepackage[T1]{fontenc}
\usepackage{tcolorbox}
\usepackage{float}

\usepackage[utf8]{inputenc}
\usepackage{booktabs}

\usepackage{microtype}

\usepackage{inconsolata}

\usepackage{graphicx}

\title{Why Do LLM-based Web Agents Fail? A Hierarchical Planning Perspective}

\author{Mohamed Aghzal, Gregory J. Stein, Ziyu Yao \\
         Department of Computer Science, George Mason University \\
         \texttt{\{maghzal,gjstein,ziyuyao\}@gmu.edu}}

\begin{document}
\maketitle

\begin{abstract}

Large language model (LLM) web agents are increasingly used for web navigation but remain far from human reliability on realistic, long-horizon tasks. Existing evaluations focus primarily on end-to-end success, offering limited insight into where failures arise. We propose a hierarchical planning framework to analyze web agents across three layers (i.e., high-level planning, low-level execution, and replanning), enabling process-based evaluation of reasoning, grounding, and recovery. Our experiments show that structured Planning Domain Definition Language (PDDL) plans produce more concise and goal-directed strategies than natural language (NL) plans, but low-level execution remains the dominant bottleneck. These results indicate that improving perceptual grounding and adaptive control, not only high-level reasoning, is critical for achieving human-level reliability. This hierarchical perspective provides a principled foundation for diagnosing and advancing LLM web agents.  

\end{abstract}

\section{Introduction}

\begin{figure*}
    \centering
    \includegraphics[width=\textwidth]{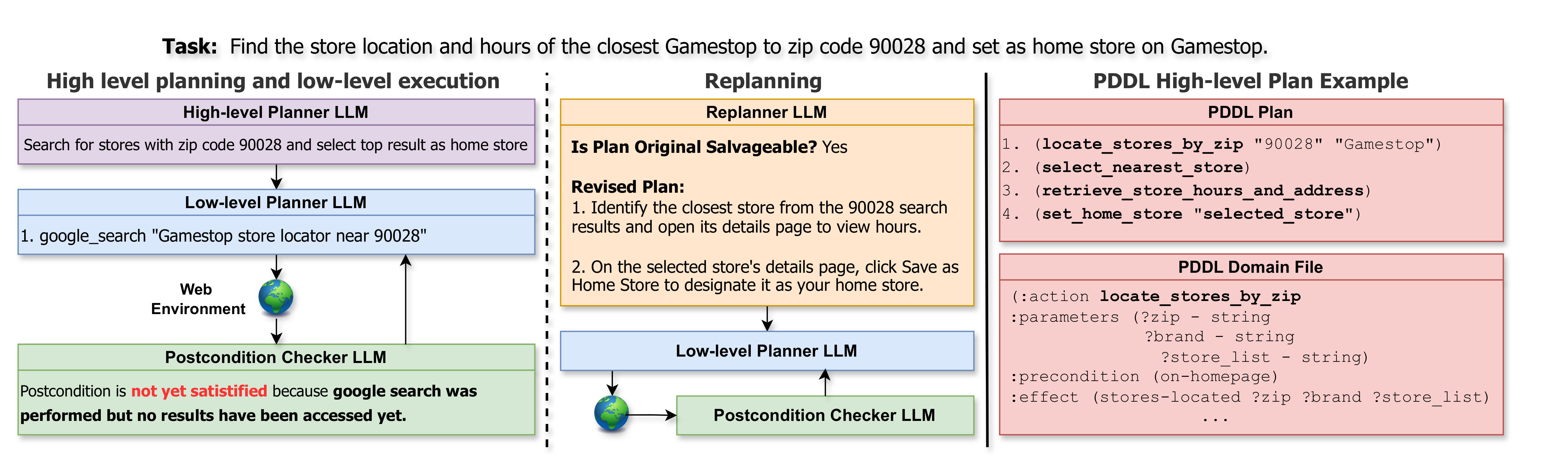}
    \vspace{-0.25in}
    \caption{\small\textbf{Overview of the hierarchical planning evaluation framework we propose.} The pipeline consists of 3 stages:  \textcolor{mypurple}{\textbf{1) \uline{High-level Planning:}}} The LLM proposes high-level subgoals,  \textcolor{myblue}{\uline{\textbf{2) Low-level Execution:}}} each high-level subgoal is translated into a set of low-level actions, a \textcolor{mygreen}{\uline{postcondition checker}} verifies whether the low-level actions lead to successful completion of the subgoal. If the subgoal fails after multiple iterations, then  \textcolor{myorange}{\uline{\textbf{3) Replanning}}} is triggered. In addition to natural language, we explore a structured representation (\textcolor{mypink}{\uline{PDDL}}) for high-level planning.}
    \vspace{-0.25in}
    \label{fig:overview}
\end{figure*}

Large language model (LLM) web agents have demonstrated substantial progress on tasks involving web browsing and multi-step online workflows \cite{zhu2023webarena,pan2024webcanvas,anupam2025browserarenaevaluatingllmagents}. These systems hold promise to automate everyday online interactions from managing e-commerce transactions to navigating corporate portals. However, despite steady improvements in reasoning and instruction-following capabilities \cite{wei2022chain,chen2023program,learning_to_reason_with_llms_2024}, current agents fall far short of human reliability on real, dynamic websites \cite{pan2024webcanvas,mialon2023gaia,zhou2023webarena,xue2025illusionprogressassessingcurrent}. 

A central limitation of existing work is its reliance on coarse, end-to-end success metrics. While these metrics quantify overall performance, they reveal little about \emph{where} failures originate: incorrect task interpretation, flawed high-level strategy, poor grounding of plans into concrete UI actions, or inadequate recovery when the environment diverges from expectations. 


To address this gap, we study approaches to systematically perform fine-grained analysis of LLM-based web agents. Current LLM-based web agents are designed under a variety of frameworks, e.g., with customized components \cite{cai2025large, chae2025web} or through direct prompting \cite{yang2025agentoccam}. However, regardless of their specific framework architecture, these agents all demand three core capabilities from the backend LLMs, including: (1) \emph{high-level planning}, i.e., the capability for LLMs to decompose a long-trajectory task into multiple subgoals (e.g., ``filter by price, sort results, then select the top item''), (2) \emph{low-level execution}, i.e., the capability of LLMs to realize the high-level subgoal through a sequence of executable actions (e.g. UI-level clicks and scrolls), and (3) \emph{replanning}, i.e., the capability of LLMs to adapt their approach when the environment does not align with their expectations (e.g. pages load unexpectedly or search results differ from expectations). {We discuss existing research improving web agents along these capabilities in Section~\ref{sec: related_work}.} This three-layer capability structure is not only considered fundamental in automated planning \cite{ghallab2004automated,erol1994htn,nau2003shop2,kaelbling2011tamp} but also closely mirrors human reasoning \cite{correa2025exploring, balaguer2016neural, eckstein2021hierarchical}. Accordingly, in this work, we focus on analyzing these three capabilities of state-of-the-art LLMs, grounded in web agent tasks. Specifically, we propose a hierarchical planning-based evaluation framework, which naturally structures an LLM-based web agent into these three layers and is thus able to scrutinize LLMs along these capability dimensions in web agent applications.

To enable this analysis, we extend the Mind2Web-Live benchmark \cite{pan2024webcanvas} with human-aligned high-level plans derived from expert key-node annotations. Using this enriched benchmark, we conduct a systematic decomposition of LLM web agent performance across all three identified layers of hierarchical planning on three base models, including \texttt{gpt-5-nano} \cite{gpt-5_2025}, \texttt{claude-haiku-4.5} \cite{claude_haiku}, and \texttt{gemini-flash-2.5} \cite{gemini2025}. 

Our analysis reveals three key findings: (1) {LLMs generate more effective high-level plans when prompted using Planning Domain Definition Language (PDDL)};
(2)  LLM performance is limited by perceptual grounding and control, highlighting current weaknesses in LLM-driven low-level execution; and (3) a single round of replanning informed by exploratory feedback substantially improves both subgoal completion and overall task success, demonstrating that replanning is an effective mechanism for improving LLM reliability. 
{These findings are shared by all three LLMs experimented in our work, although \texttt{gpt-5-nano} exhibits superior performance than the other two models and \texttt{gemini-flash-2.5} particularly struggles with low-level execution.}
{We release our evaluation framework and data for future research.\footnote{\url{https://github.com/Ziyu-Yao-NLP-Lab/llm-hierarchical-web-agents}}}

\section{Preliminaries: Hierarchical Planning-based Web Agents}
\label{sec:hp-definition}
Figure \ref{fig:overview} presents an overview of the hierarchical planning framework for LLM-based web agents, which we use for LLM evaluation in Section~\ref{sec:results}.
We describe each component in this section.

\subsection{High-level Planning}

At the top level, the agent must derive a sequence of high-level \emph{subgoals} from a natural language instruction \( I \).  
These subgoals represent abstract, meaningful steps that collectively describe how to achieve the overall objective.  
We denote this high-level plan as \( P = [g_1, g_2, \dots, g_n] \), where each \( g_i \) is a subgoal that can be grounded into a sequence of low-level executable actions.

\subsection{Low-level Execution}

Given a high-level plan \( P = [g_1, \dots, g_n] \), the agent must instantiate each subgoal \( g_i \) into an actionable sequence of low-level web interactions.  
Each low-level step corresponds to choosing an action \( a_t \in \mathcal{A} \) based on the current observation \( o_t \in \mathcal{O} \), producing a trajectory  
\( \tau_i = (o_t, a_t, o_{t+1}, \dots, o_{t+k}) \) that attempts to satisfy \( g_i \).  
Actions are drawn from a discrete space defined by the web environment (e.g., clicking elements, performing google search, etc.).

\subsection{Postcondition Checking}

For the agent to determine whether a subgoal \( g_i \) has been successfully achieved, it must evaluate the outcome of its executed actions against the intended postconditions of the subgoal.  
This process serves as the feedback signal that determines whether to proceed to the next subgoal, retry the current one, or trigger replanning. After executing a trajectory \( \tau_i \) associated with subgoal \( g_i \), the agent observes a state \( s' \) (and \( \tau_i \)) and produces the evaluation
\( \Phi \)
where \( \Phi(g_i, s') = 1 \) if and only if the effects of \( g_i \) are satisfied in \( s' \).  We implement postcondition checking using an LLM-as-judge, which assesses whether the agent's recent actions have accomplished the intended intermediate goal. 

\subsection{Replanning}

Realistic web environments are noisy and unpredictable, and prior assumptions about the dynamics and structure of websites may prove invalid.  
Therefore, hierarchical agents must incorporate a \emph{replanning} mechanism to detect and recover from failures.  
If a subgoal fails or leads to a dead end, the agent can revise its plan in one of two ways: (1) \emph{Local adjustment:} continue planning from the last successful subgoal in the plan, and (2) \emph{Global replanning:} discard the existing plan and generate a new sequence of subgoals conditioned on the current context. In order to achieve this, the LLM is prompted to first choose whether the original high-level plan is salvageable based on the current state and history; if the agent decides that the plan is not recoverable, then it proposes a new plan from scratch, otherwise it proposes the next high-level subgoals to take from the current state. Analysis in this work is limited to one replanning round.

\section{Hierarchical Evaluation of LLMs for Web Agent Applications}
{Leveraging the hierarchical planning-based agent framework, we}
evaluate LLMs on three fundamental skills, i.e., \emph{high-level planning}, \emph{low-level execution}, and \emph{replanning}, that enable them to eventually support web agent applications.

\subsection{Why Evaluating via a Hierarchical Planning Perspective?}


\noindent\textbf{A structured basis for analysis.} Long-horizon tasks naturally decompose into high-level goals, low-level actions, and replanning when progress diverges from expectations. This makes high-level planning, low-level execution and replanning three fundamental skills for web agents. By explicitly scrutinizing web agents along these three capability dimensions, hierarchical decomposition enables analysis beyond task success, revealing how design choices affect LLM performance across each of these skills. Additionally, outcome-based evaluation alone is insufficient, as aggregate success rates obscure the sources of agent failure by conflating planning errors, execution mistakes, and ineffective feedback use; a hierarchical evaluation framework makes these failure modes explicit, enabling attribution of errors to specific core capabilities.


\noindent\textbf{Human-aligned reasoning.} Humans solve complex tasks hierarchically, forming abstract strategies, executing concrete actions, and revising plans when necessary \cite{balaguer2016neural,eckstein2021hierarchical,correa2025exploring}. Evaluating web agents through a similar structure aligns it with models of human problem-solving and supports comparisons between human and agent reasoning.


In our analysis, we aim to highlight a set of our research questions with respect to each of the dimensions described in Section \ref{sec:hp-definition}.

\subsection{High-level Planning}

\subsubsection{Research Questions}

\textbf{RQ1:} \textbf{\emph{Do LLM agents generate high-level plans that align with human-authored subgoals?}} We analyze the alignment between the human-annotated high-level plans and the LLM-generated ones. Namely, we evaluate the degree to which LLM-generated plans reflect the structure and intent of human-authored high-level plans.

\noindent\textbf{RQ2:} \textbf{\emph{Do structured representations (e.g., PDDL) improve alignment with human high-level plans compared to natural language?}} We look into whether inducing structure and symbolic reasoning through formal representations such as PDDL yields better alignment with human plans.

\noindent\textbf{RQ3:} \textbf{\emph{How executable are the high-level goals produced by each representation?}} Finally, we compare goals produced by NL as well as PDDL against a human oracle in terms of executability. In other words, we compare how well a low-level executor LLM is able to perform each of the high-level subgoals on the environment. 

\subsubsection{Evaluation Metrics}

\begin{table*}[tb]
\centering
\setlength{\tabcolsep}{10pt}
\renewcommand{\arraystretch}{1.3}
\resizebox{0.95\linewidth}{!}{%

\small
\begin{tabular}{p{0.28\linewidth} p{0.42\linewidth} p{0.26\linewidth}}
\toprule
\textbf{Human Step} & \textbf{NL (LLM-generated) Step} & \textbf{PDDL (LLM-generated) Step} \\
\toprule

Set the search criteria to include Audi cars made between 1992 and 2015. & \textcolor{purple}{Apply a year filter to restrict results to model years before 2015 (i.e., 2014 and earlier).} & \textcolor{purple}{\texttt{(filter\_by\_year "before 2015")}} \\
\hline
Go to the E-Gift card purchase page. &
\textcolor{darkgreen}{Navigate to the e-gift card section.} &
\textcolor{darkgreen}{\texttt{(open\_gift\_card\_section)}} \\

\hline

Add the highest rated activity to your wish list. &
\textcolor{blue}{From the results, identify the card with the highest rating and open its details by clicking on that card.                                 
Add the identified experience to your wishlist by clicking the Save to Wishlist (or Add to Wishlist) control on the details page.} &
\textcolor{blue}{\texttt{(select\_top\_result)};
\texttt{(add\_to\_wishlist)}} \\
\midrule
\multicolumn{3}{l}{\textbf{\textcolor{graytext}{Unmatched LLM steps (reference human plan shown below):}}} \\
\cmidrule(lr){2-3}

\textcolor{graytext}{1. Navigate to the 'The Legend of Zelda: Breath of the Wild' game page.} & 
\textcolor{graytext}{If no suitable walkthrough is found, perform an alternate search for Breath of the Wild guides (in Nintendo or Games sections)} & 
\textcolor{graytext}{\texttt{(verify\_walkthrough\_loaded)}} \\

\textcolor{graytext}{2. Access the walkthrough section for 'The Legend of Zelda: Breath of the Wild'} & & \\

\bottomrule
\end{tabular}
}
\vspace{-0.1in}
\caption{
\small Examples of annotated NL and PDDL steps with reference to human steps.
\textcolor{blue}{Blue} = Decomposed alignment,
\textcolor{darkgreen}{Green} = Perfect Match,
\textcolor{purple}{Purple} = Partial alignment,
\textcolor{graytext}{Gray} = Unmatched or extra steps.
}
\label{tab:example-plans}
\vspace{-0.2in}
\end{table*}

We measure alignment between LLM-generated and human-annotated high-level plans using the metrics illustrated in Figure~\ref{fig:alignment-tree} (Appendix~\ref{app:alignment-trees}). (1) \emph{Perfect Match} is the proportion of human-authored steps with a corresponding action or intent in the LLM-generated plan. (2) \emph{Partial Rate} measures cases where an LLM step captures only part of a human step (e.g., ``select an item'' vs. ``select the first item''). (3) \emph{Missing Rate} denotes human steps absent from the LLM plan, reflecting gaps in task representation. (4) \emph{Decomposed Rate} captures instances where a single human step is split into multiple finer-grained LLM steps. (5) \emph{Unmatched Rate} measures LLM steps with no corresponding human step, indicating redundant or irrelevant actions. (6) The \emph{Matched Rate}, measures the proportion of LLM-generated steps that can be aligned to at least one human-authored step. Example annotations are shown in Table~\ref{tab:example-plans}. These metrics are computed by providing an LLM with both the oracle and generated plans and instructing it to assign each oracle step to one of the first six categories. We additionally compare executability using completion rates (Section~\ref{sec:low-level-metrics}).

\subsection{Low-level Execution}

\subsubsection{Research Questions}

\noindent\textbf{RQ4:} \textbf{\emph{Given accurate high-level plans, how reliably can LLM agents execute the required low-level actions to achieve subgoals?}}  
We examine the agent's ability to ground the human subgoals into a correct sequence of concrete web interactions. By fixing the high-level plan, we isolate low-level execution and assess the agent's execution capabilities.

\noindent\textbf{RQ5:} \textbf{\emph{What are the main failure modes in low-level execution, and how do they affect overall task performance?}}  
We perform an error analysis to categorize frequent execution failures in order to shed light on the limits of LLMs' in converting high-level subgoals into low-level actions.

\subsubsection{Evaluation Metrics}\label{sec:low-level-metrics}

We evaluate low-level execution using the following metrics: (1) \emph{Subgoal Completion Rate}, defined as the proportion of subgoals from the high-level plan that are successfully completed. Each subgoal is determined by an LLM-based postcondition checker using the observed state and action history; (2) \emph{Plan Completion Rate}, which measures the fraction of high-level plans for which all constituent subgoals are completed; (3) \emph{Plan Efficiency}, computed as the average number of actions required to complete tasks, and (4) \emph{Task Success Rate}, measured by an LLM-based judge that determines whether the original instruction is satisfied given the full action history and final webpage.

\subsection{Replanning}

\subsubsection{Research Questions}

    \noindent\textbf{RQ6:} \textbf{\emph{Does replanning improve alignment with human-authored plans in the presence of partial or incorrect initial plans?}} We assess how often replanning leads to alignment with the human-annotated high-level plans.
        
	\noindent\textbf{RQ7:} \textbf{\emph{Can agents effectively revise their high-level plans?}} To evaluate replanning capabilities, we analyze whether the revised high-level plans that agents produce through replanning lead to better overall task success. 

\subsubsection{Evaluation Metrics}

We assess replanning by looking at how the performance is affected after replanning, both in terms of planning and execution. Mainly, we look at the alignment metrics of the high-level plans after replanning as well as the overall success rate. 

\subsection{LLM-as-Judge Evaluation}

As argued by previous work \cite{xue2025illusionprogressassessingcurrent,zhang2023cumulative,pan2024autonomous,he-etal-2024-webvoyager,yu2025aisjudgeaisrise,gou2025mindweb2}, LLM-as-Judge approaches, where a language model serves as an automatic evaluator by assessing trajectories or final webpage states relative to the task instruction can leverage the model's semantic understanding to provide flexible, context-aware judgments. This is the approach we adopt in our work as well for both postcondition checking and overall evaluation. We assess the reliability of this evaluation by conducting manual verification for a subset of the outputs, which can be found in Appendix \ref{app:judge-reliability}.

\section{Experimental Setup}
We provide additional details about the experimental setup, including prompts and more detailed examples in Appendix \ref{app:experimental-design}.

\subsection{Dataset Construction}

In this work, we use the test set of the Mind2Web-Live benchmark \cite{pan2024webcanvas}, which consists of 104 instances of web tasks performed on live websites. It provides expert-annotated key nodes, corresponding to intermediate subgoals identified as essential for completing a web task. We treat these annotations as gold-standard high-level subgoals. We prompt \texttt{gpt-5-nano} \cite{gpt-5_2025} to convert the JSON key-node representation into a coherent sequence of abstract actions, yielding a human-readable high-level plan that captures the expert’s intended task progression without low-level details. We manually verify the generated plans for (1) \emph{correctness}, ensuring fidelity to the original key nodes without omissions, hallucinations, or semantic reordering, and (2) \emph{clarity and granularity}, ensuring each step is a clear, meaningful high-level subgoal. We find that while step content is generally correct, some steps are poorly phrased (e.g. framed as an evaluation rather than a step), incomplete (e.g. part of the requirements is missing), or overly specific (e.g. directly navigating to a link); we manually correct these issues.

\subsection{Models}

We use \texttt{gpt-5-nano} \cite{gpt-5_2025} as the primary model for all stages of our pipeline, including action generation and evaluation, due to its strong balance between performance and computational efficiency. To assess generality, we additionally evaluate two other models that are similar in terms of cost and performance \texttt{gemini-flash-2.5} \cite{gemini2025} and \texttt{claude-haiku-4.5} \cite{claude_haiku}. We note that evaluation is still done using \texttt{gpt-5-nano} even for these other models to ensure stability. 

Across all experiments, agents observe the environment through a text-based DOM representation of the current web page, including visible elements, their attributes, and structural relationships. We evaluate three action settings that differ in both the action space and the representation of actions. In the \emph{Expanded} setting, adapted from \citet{pan2024webcanvas}, the agent operates over an expressive action space including actions such as \texttt{google\_search}, \texttt{goto}, \texttt{click}, \texttt{fill\_form}, and \texttt{get\_final\_answer}, which abstracts over low-level UI interactions and is more expressive than the primitive Mind2Web action set \cite{mind2web2023}. We contrast this with a restricted \emph{primitive} action space consisting only of low-level UI actions (\texttt{click}, \texttt{type}, \texttt{select}, and \texttt{hover}), evaluated under two different representations. In the \emph{Action Object} setting, the model generates the full action specification, including the action type and the target element. In the \emph{Action ID} setting, the model instead selects from a predefined list of valid low-level actions enumerated for the current page and outputs only the corresponding action identifier. 

\subsection{High-level Representations}



The choice of high-level representation is critical in hierarchical planning. Natural language is flexible but lacks explicit planning structure, often producing  (as we demonstrate below) \emph{underspecified} plans (e.g., missing implicit requirements) or \emph{overspecified} plans (e.g., low-level assumptions that may not hold), and provides no mechanism to enforce plan structure. These limitations motivate the use of structured formal representations such as PDDL. Accordingly, we study two high-level planning representations. These include \emph{1) Natural Language (NL):} The LLM directly produces subgoals as free-form text, and \emph{2) PDDL} \cite{ghallab1998pddl}, where subgoals are expressed in a structured form that enforces explicit preconditions and effects. 

\section{Results and Analysis}\label{sec:results}
We discuss the results for \texttt{gpt-5-nano} in this section. We present further analyses of additional models and more examples in Appendix \ref{app:aditional-analysis}.

\subsection{High-level Planning}\label{sec:high-level}

\begin{table}[tb]
    \centering
    
    \resizebox{\columnwidth}{!}{%
    \begin{tabular}{lcccc}
        \toprule
        & \multicolumn{2}{c}{\textbf{Before Replanning}} 
        & \multicolumn{2}{c}{\textbf{After Replanning}} \\
        \cline{2-5}
        & \textbf{NL} & \textbf{PDDL} & \textbf{NL} & \textbf{PDDL} \\
        \midrule
        \textbf{Perfect Match} & 60.6 & 67.7 & 56.1 & 59.0 \\
        \textbf{Partial}       & 5.7  & 7.4  & 6.1  & 6.9  \\
        \textbf{Missing}       & 4.2  & 2.2  & 4.0  & 14.5 \\
        \textbf{Decomposed}    & 29.5 & 22.7 & 33.8 & 19.6 \\
        \midrule
        \textbf{Unmatched}     & 29.4 & 15.4 & 35.0 & 15.4 \\
        \textbf{Matched}       & 70.6 & 84.6 & 65.0 & 84.6 \\
        \bottomrule
    \end{tabular}%
    }
    \vspace{-0.1in}
    \caption{\small Alignment (\%) between human and LLM-generated high-level plans.}
    \label{tab:high-level}
    \vspace{-0.25in}
\end{table}

\noindent\textbf{RQ1: Do LLM agents generate high-level plans that align with human-authored subgoals?}
Table \ref{tab:high-level} shows the alignment results of the LLM  high-level plans with the human ones.
\emph{LLM-generated plans exhibit only partial alignment with human-authored subgoals}. As shown in Table~\ref{tab:high-level}, 60.6\% of steps in natural language (NL) plans and 67.7\% in PDDL-based plans correspond directly to human-annotated ones, suggesting that while LLMs capture the overall task structure, they often fail to match the abstraction level of human reasoning. Most deviations arise from over-specification rather than omission: 29.5\% of NL steps are decomposed compared to only 4.2\% missing, meaning models tend to expand single human steps into multiple finer-grained subgoals. In addition, 29.4\% of NL and 15.4\% of PDDL steps are unmatched to any human subgoal, revealing redundant or spurious actions. These findings show that LLMs generally understand task intent but generate overly detailed, noisy plans that diverge from human abstractions.

\noindent\textbf{RQ2: Does the structured PDDL representation improve alignment with human plans compared to natural language?}
\emph{The PDDL representation
yields more concise, goal-directed plans}. Compared to NL plans, PDDL-based plans achieve higher step existence (67.7\% vs.\ 60.6\%) and lower missing rates (2.2\% vs.\ 4.2\%), 
while also reducing decomposed (22.7\% vs.\ 29.5\%) and unmatched (15.4\% vs.\ 29.4\%) steps. This suggests that \emph{symbolic constraints encourage clearer, more efficient planning} by reducing verbosity and irrelevant elaborations. Consequently, the higher matched rate (84.6\% vs. 70.6\%) demonstrates that structured formulations better capture the core logic of human high-level reasoning and provide a more stable foundation for subsequent execution. We also observe from Table \ref{tab:efficiency} that using \emph{PDDL yields more compact plans and shorter low-level sequences}.

\begin{figure}

    \centering
    \includegraphics[width=\linewidth]{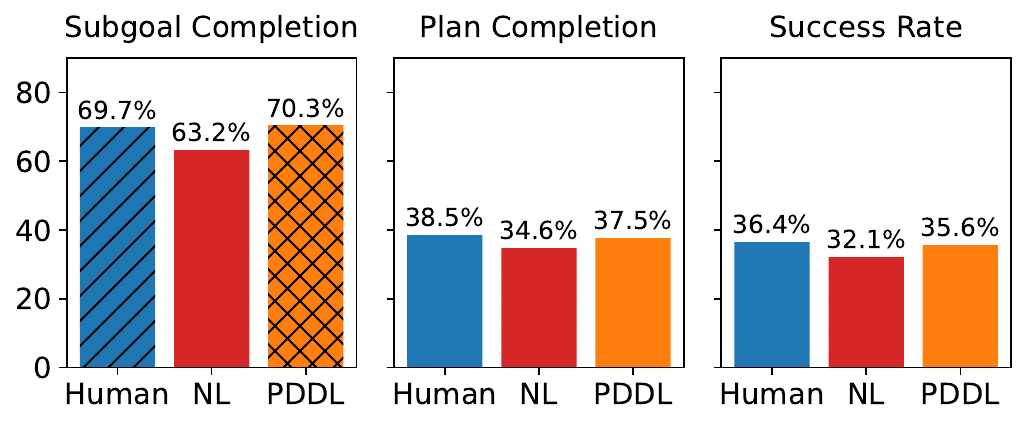}
    \vspace{-0.3in}
    \caption{Execution results of different representations}
    \label{fig:execution_results}
    \vspace{-0.15in}
\end{figure}

\noindent\textbf{RQ3: How executable are the high-level goals produced by each representation?}
Human-authored plans are more effective than NL plans when executed by LLM web agents using the expanded action space (Figure~\ref{fig:execution_results}). \emph{PDDL plans largely bridge this gap}, performing only slightly worse than humans. Additionally, while 68\% (NL) and 73\% (PDDL) of decompositions render previously non-executable subgoals executable, decomposing compact subgoals can hurt executability.

\begin{table}[tb]
    \centering
    \small
    \begin{tabular}{lccc}
        \toprule
        & \textbf{Human} & \textbf{NL} & \textbf{PDDL} \\
        \midrule
        \textbf{\# High-level subgoals} & 3.04 & 5.51 & 4.39 \\
        \textbf{\# Low-level actions} & 7.08 & 11.94 & 8.89 \\
        \bottomrule
    \end{tabular}
    \vspace{-0.1in}
    \caption{ Avg. length of high-level plans and low-level actions for each representation.}
    \label{tab:efficiency}
    \vspace{-0.26in}
\end{table}

\begin{table}[t]
    \centering
    \resizebox{\columnwidth}{!}{
        \begin{tabular}{lccc}
            \toprule
            \textbf{Metric} & \textbf{Action Object} & \textbf{Action ID} & \textbf{Expanded} \\
            \midrule
            Hallucination Rate & 34.0 & 2.0 & 3.0 \\
            \bottomrule
        \end{tabular}
    }
                \vspace{-0.1in}

    \caption{\small Hallucination (invalid action) rates across different action representations (\%).}
    \label{tab:hallucination-rate}
    \vspace{-0.35in}
\end{table}

\subsection{Low-level Execution}\label{sec:low-level}
\begin{table}[t]
    \centering
    \resizebox{0.9\columnwidth}{!}{
        \begin{tabular}{lcl}
            \toprule
            \textbf{Failure Mode} & \textbf{Rate (\%)} & \textbf{Base Category} \\
            \midrule
            Repetitions & 10.4 & of failures \\
            Hallucinated links & 32.0 & of \texttt{goto} actions \\
            Redundant actions & 34.2 & of all actions \\
            Out-of-domain links & 16.7 & of all actions \\
            \bottomrule
        \end{tabular}
            \vspace{-0.25in}
    }
    \caption{\small {Failure modes across different categories.}
    Each rate is computed relative to its own base set. Illustrated examples of these failures can be found in Appendix \ref{app:low-level-failures}. }
    \label{tab:failure-modes}
        \vspace{-0.18in}

\end{table}

\noindent\textbf{RQ4: Given accurate high-level plans, how reliably can LLM agents execute the required low-level actions to achieve subgoals?}
Figure~\ref{fig:execution_results} shows that \emph{even when provided with human-authored high-level plans, LLM-based executors struggle to consistently translate subgoals into correct low-level actions}.
 The executor LLM achieves only a 38.5\% plan completion rate and a 36.4\% final success rate. This suggests that the model struggles to ground abstract subgoals into executable actions, making low-level execution a key bottleneck in web tasks. 

\noindent\textbf{RQ5: What are the main failure modes in low-level execution, and how do they affect overall task performance?} 
In addition to the metrics computed in section \ref{sec:low-level-metrics}, we also compute the following metrics to quantify observed errors; (1) \emph{Action Validity}, defined as the proportion of proposed actions that target valid DOM elements, with invalid or hallucinated actions indicating perception failures; (2) \emph{Hallucinated Links}, measured as the percentage of actions that lead to invalid links, judged by \texttt{gpt-5-nano} based on whether any information is displayed on the page; (3) \emph{Action Repetition Rate}, which tracks the proportion of failures due to stagnant repetition; (4) \emph{Redundancy Rate}, defined as the rate of valid actions that do not produce any state changes, indicating poor understanding of action effects and dynamics.

\emph{Different action spaces fail for different reasons.}
As shown in Table \ref{tab:hallucination-rate}, invalid actions are common with full action-object prediction and much lower with action-ID selection and the expanded action space. Full action objects prompt invented actions under uncertainty, while ID selection defaults to arbitrary choices, and expanded actions often fall back to \texttt{google\_search} or \texttt{goto} when stuck. 
Although this can escape local dead ends, it frequently causes off-task behavior: direct links may not exist, and \texttt{google\_search} can redirect outside the target site (e.g., retrieving the information from a different website). As can be observed in Table \ref{tab:failure-modes}, over 16\% of actions occur outside the required domain, and 32\% of \texttt{goto} actions lead to non-existent links.

\noindent\emph{LLM executors frequently get stuck in repetitive or redundant behaviors.} We observe that low-level executors frequently repeat failed actions despite environmental feedback. 10.4\% of failures arise from the agent executing the same action multiple times (more than 3 times), and 34\% of actions produce no change in the DOM. This suggests limited state understanding even after exploration.

\subsection{Replanning}

\begin{figure}[t]
    \centering
    \includegraphics[width=\linewidth]{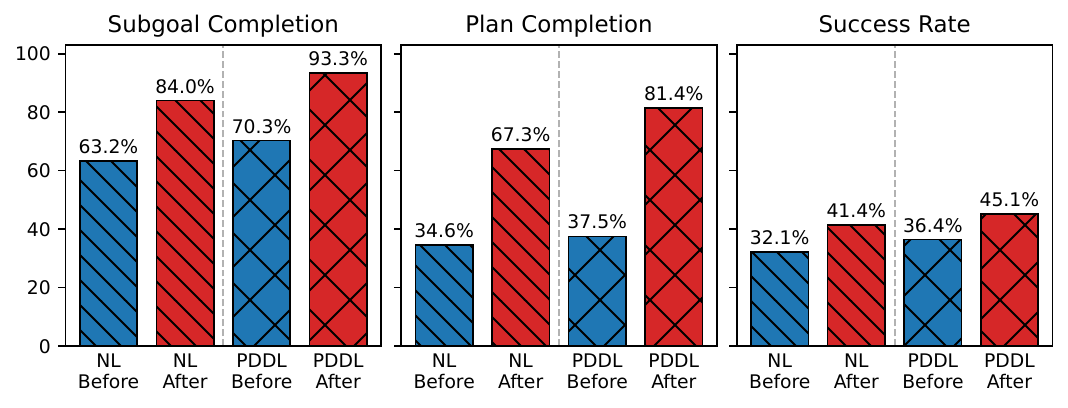}
    \vspace{-0.25in}
    \caption{Performance with and without replanning}
    \label{fig:replanning}
    \vspace{-0.25in}
\end{figure}

\paragraph{RQ6:} \textbf{{Does replanning improve alignment with task goals in the presence of partial or incorrect initial plans?}} As can be seen in Table \ref{tab:high-level}, \emph{replanning slightly worsens alignment for NL plans}: perfect matches drop from 60.6\% to 56.1\%, and added steps increase fragmentation, reducing the matched proportion from 79.6\% to 65.0\%. Rather than refining plans, NL representations drift toward over-elaboration after exploration. \emph{PDDL plans also degrade}, but differently. Missing steps rise sharply (2.2\% to 14.5\%) and perfect matches fall (67.7\% to 59.0\%), while unmatched steps remain stable and over-decomposition slightly decreases. Nevertheless, PDDL is more stable than NL, with a fixed unmatched rate (15.4\% vs. 35.0\%) and a higher matched rate (84.6\% vs. 65.0\%).

\paragraph{RQ7:} \textbf{{Can agents effectively revise their high-level plans?}} Figure \ref{fig:replanning} shows the overall success rate and subgoal completion rate before and after replanning for both NL and PDDL; \emph{we can see that the models can improve substantially after one round of replanning}. The high-level subgoals produced upon replanning, while less aligned with humans, yield a significantly higher subgoal completion rate, implying that planning after exploring the environments leads to more executable plans.  

\subsection{How Do Different LLMs Perform?}

{With our evaluation framework, we are able to compare different LLMs along the three core capabilities in web agent tasks.}
We present results for two other models, \texttt{claude-haiku-4.5} and \texttt{gemini-flash-2.5}, in Appendix \ref{app:other-models}, and highlight our main observations in this section.

\noindent\textbf{\texttt{gpt-5-nano} exhibits superior performance.} Both models perform significantly worse than \texttt{gpt-5-nano} (29.2\% success rate for \texttt{claude-haiku-4.5} and 17.3\% for \texttt{gemini-flash-2.5} with human plans vs 36.4\% with \texttt{gpt-5-nano}).

\noindent\textbf{\texttt{gemini-flash-2.5} yields more compact plans but struggles with executing them} The high-level plans produced by \texttt{gemini-flash-2.5} are significantly more compact than the ones produced by other models. However, it exhibits the strongest limitations in low-level execution, achieving the lowest subgoal completion, plan completion, and success rates. It also shows the highest rate of redundant actions (41.2\% of all actions, compared to \texttt{claude-haiku-4.5}'s 25.6\% and \texttt {gpt-5-nano}'s 34.2\%), indicating a poor understanding of the environment. This limitation is further reflected in the model's minimal gains from replanning.

\noindent\textbf{\texttt{claude-haiku-4.5} does not rely on link navigation as much as the other two models} \texttt{claude-haiku-4.5} does not instantiate \texttt{goto} and \texttt{google\_search} actions as often as the other models. It also achieves the lowest rate of hallucinated links. However, it achieves the highest rate of repetitions, indicating a lower ability to use feedback.

\section{Recommendations for Web Agent Design from a Hierarchical Perspective}
Our framework highlights a few directions for improving web agents from a hierarchical perspective.

\noindent\textbf{1. Treat planning and execution as separate, complementary skills.}
The discrepancy in performance between high-level planning and low-level execution that we observe suggest that planning and execution require distinct capabilities and evaluation schemes, (e.g. evaluating how well abstracted a high-level plan is vs. how well a low-level executor responds to observations). We, thus, advocate for agents that explicitly separate these components: high-level modules that produce well-abstracted subgoals, and low-level controllers for robust grounding. Structured representations such as PDDL are well-suited for the former, while specialized tools may be required for the latter.

\noindent\textbf{2. Design better methods for perceptual grounding.}
Our error analysis in Section~\ref{sec:low-level} shows that mis-grounding (e.g., hallucinated or visually similar elements) and poor state tracking (e.g., repeated actions and redundant navigation) are primary failure modes. Addressing these issues will require richer grounding and state representations that help models track progress and use feedback effectively.

\noindent\textbf{3. Design action spaces that respect uncertainty in the world model.}
Different action representations induce different failure modes: predicting full action–object pairs encourages hallucinated UI operations, while action-ID or expanded actions reduce invalid actions but often reflect uncertainty as random errors. We argue that action spaces should allow the executor to explicitly signal uncertainty or world-model mismatch (e.g., clarification or replanning actions) rather than forcing a concrete UI action at every step, better aligning execution with the planner's internal world model.

\noindent\textbf{4. Make fine-grained process-based evaluation a standard.} Beyond end-to-end success rates, process-based evaluation across planning, execution, and replanning enables more precise diagnosis of agent failures and targeted comparison across architectures. Complementary tools such as mechanistic interpretability \cite{rai2025practicalreviewmechanisticinterpretability} can further reveal how models represent their environments and decisions, informing systematic improvement.

\section{Related Work} \label{sec: related_work}


\noindent\textbf{Evaluation of LLM Web Agents.}
Evaluating LLM-based web agents on realistic, live websites remains challenging. Early benchmarks in offline settings rely on exact trajectory matching against ground-truth demonstrations \cite{mind2web2023}, which does not scale to dynamic web environments. As a result, recent work has explored alternative strategies, including manual human assessment on live sites \cite{zheng2023seeact}, rule-based success criteria \cite{pan2024webcanvas,zhu2023webarena}, and LLM-as-Judge approaches that assess trajectories or final states relative to the task instruction \cite{xue2025illusionprogressassessingcurrent,zhang2023cumulative,pan2024autonomous,he-etal-2024-webvoyager,yu2025aisjudgeaisrise,gou2025mindweb2}. Our work advocates evaluation that disentangles high-level planning, low-level execution, and replanning to understand why web agents fail.


\noindent\textbf{High-level Planning with LLMs} Recent work shows that LLMs can generate high-level plans. Many approaches prompt LLMs to produce structured representations (e.g. PDDL domains or temporal logic specifications) that classical planners can consume \cite{silver2022pddl, xie2023translating, guan2023leveraging, chen2024autotamp, xu2024learningltl}. In the web domain, LLM-based systems decompose user instructions into abstract subgoals that guide execution \cite{kim2024llmcompiler, erdogan2025planandact, yang2025selfguided}. Our work extends this line by evaluating NL and PDDL. To our knowledge, this is the first work to incorporate formal planning representations into web agents.

\noindent\textbf{Improving LLMs' Low-level Execution} Grounding high-level skills into low-level actions is a key milestone for web agents \cite{zheng2024seeact}. Several work thus aims to improve the reliability of LLM-based executors in web environments. One common approach simplifies webpage observations to better align with LLM representations \cite{zhu2023webarena, yang2025agentoccam}. Other methods abstract low-level UI interactions into website-specific tools, improving robustness by avoiding fragile element-level manipulation \cite{kim2024llmcompiler, prabhu2025walt}. In contrast, some approaches rely on training to learn actions directly from trajectories \cite{qi2025webrl, chae2025web}. We propose metrics that evaluate low-level execution with respect to explicit high-level goals while maintaining a separation between the two.

\noindent\textbf{Replanning and Self-Refinement} Early agent prompting frameworks interleave reasoning with actions in a loop adapting to new observations at each step \cite{yao2023react, shinn2023reflexion}, showing the potential of LLM planning in partially observed environments. This inspired several approaches that leverage environment feedback to perform replanning for web agent applications \cite{yang2025webdartdynamicdecompositionreplanning, erdogan2025planandact, zhang2025reflectingvoicescoadaptivedualstrategy, patel2024largelanguagemodelsselfimprove, yang2025selfguided}.
\section{Conclusion}
We evaluated LLMs on web agent tasks from a hierarchical planning perspective, analyzing high-level planning, low-level execution, and replanning. We compared NL and PDDL representations and found that structured plans can be helpful, while execution remains the main bottleneck. This evaluation framework offers a principled basis for diagnosing web agent failures and guiding future research.

\section*{Limitations}


While our hierarchical evaluation framework aims to evaluate LLMs on the three distinct levels in generality, we only experiment with a limited set of high-level representations, action spaces, and agent configurations, and do not consider multimodal settings. These directions are promising avenues for future work. However, our focus in this paper is on establishing the feasibility and diagnostic value of hierarchical evaluation for long-horizon web-based tasks. Additionally, the high-level plan evaluation requires an annotated set of reference plans for each task, making it less flexible. In this work, we assume the availability of the reference plans and focus on providing metrics that can be used for evaluating high-level plans against them. Future work can extend our work by exploring efficient annotation approaches to minimize the human effort needed.

\section*{Acknowldgments}
The work is partially supported by the PatriBot project at George Mason University and the National Science Foundation (NSF) under award No. 2418580.
This material is also based upon work supported by NSF under Grant 2232733; G. J. Stein acknowledges support from this grant and from the U.S. Army Research Laboratory (W911NF2520011).

\bibliography{custom}

\newpage

\appendix

\section{Experimental Setup}\label{app:experimental-design}

\subsection{Prompts}\label{app:prompts}

This section outlines the prompts used for the different stages of the planning pipeline we use. 

\subsubsection{High-level Planning}

\begin{tcolorbox}[width=\columnwidth,
breakable,
title={NL Prompt},
colback=grey,
colframe=black,
boxrule=0.5pt]

You are an intelligent assistant helping a user complete a task on a web page.

First, reflect step-by-step: \\
1. What is the user trying to accomplish? \\
2. What subgoals must be achieved? \\ 
3. Can each subgoal be broken into lower-level actions (click, type, etc.)? \\
4. Are these steps reusable and modular across different pages? \\

Then output your response in this strict format: \\

\textbf{Explanation}: \\
Your reasoning for proposing the high-level plan in 2-3 sentences. \\

Proposed high-level plan: \\
1. Step one  \\
2. Step two \\
...

\textbf{Rules}: \\
- Do NOT use markdown formatting. \\
- Assume the user is on the homepage of the website. Do not propose navigating to the homepage as your first action. \\
- Make sure each step is atomic, reusable, and task-driven. \\
- Focus on clarity, reusability, and structure. \\
- Make sure each step is a high-level skill that can be decomposed into low-level actions and this used to achieve a subgoal needed to achieve the task.  \\
- Do not use dummy credentials to log in. Unless explicitly asked to do so as part of the task, and credentials are provided, then you should not propose to sign in. \\
- Do not use dummy inputs. If you do not know the inputs, then you need to find a new action. \\
- You are not allowed to make up specific links. \\
- Make sure each action is atomic and reusable. \\
- Keep the actions at a high level. Do not make up any details about the parameters when you instantiate actions (i.e. if you expect that search results are to be displayed, do not “imagine” an item in the search results and use that as input. Similarly do not make up links, etc.). \\

Here is an example of a bad high-level plan for Task: ``Find the cheapest red t-shirt". \\

\textbf{High level plan}: \\

Login with example uname and password \\
Select "Nike Red T-Shirt 330"

Why this is a bad plan: \\
- This attempts to log in despite the task not asking for it. \\
- The search result is too specific and is hallucinated. \\
- If the value is unknown, the action should be abstracted or omitted. \\

A good example of a high-level plan for the same task is: \\

\textbf{High level plan}: 

Search for "red t-shirt"
Filter by price "low"
Select top result

\textbf{Task}: \{task\} \\
\textbf{Webpage}: \{web\_page\} \\
\end{tcolorbox}

\begin{tcolorbox}[width=\columnwidth,
breakable,
title={PDDL Plan Generation},
colback=grey,
colframe=black,
boxrule=0.5pt]
You are a planning assistant generating a high-level plan for a user completing a task on a web page.
\\
You need to propose a high level plan for the task. The high level plan should be a list of PDDL actions that can be performed on the website to achieve the task.
\\
Only Output the reasoning followed by the high level plan, and nothing else. Please follow the format of the example exactly. Each line should be a PDDL action instantiation. \\

\textbf{Explanation}: \\
- Why each action is needed along with a step by step explanation of how the subgoals contribute to achieving the task. \\
- Clarify what each action does.\\
- Comment on reusability across pages. \\
- Mention any assumptions made about the DOM or user context. \\
- Do not use markdown formatting. \\

\textbf{High Level Plan}: \\
1. (search\_product "iphone") \\
2. (filter\_by\_price "low") \\
...\\

\textbf{Rules}: You cannot break any of the following rules. \\
- Your plan should be consistent, and complete. It should be a high-level set of steps that if executed, will achieve the task. \\
- Do not use markdown formatting. \\
- Do not use dummy credentials to log in. Unless explicitly asked to do so as part of the task, and credentials are provided, then you should not propose to sign in. \\
- Do not use dummy inputs. If you do not know the inputs, then you need to find a new action. \\
- You are also not allowed to instantiate actions with unknown inputs (for example, (login ?username ?password) or (search ?query) are not allowed as high-level steps in the high-level plan).  \\
- You are not allowed to use links as inputs to actions. \\
- Assume the user is on the homepage of the website. Do not propose navigating to the homepage as your first action. on-homepage is a precondition that is true at the initial state of the task.
- Make sure each action is atomic and reusable. \\
- Make sure each step is a high-level skill that can be decomposed into low-level actions and thus used to achieve a subgoal needed to achieve the task. \\
- Keep the actions at a high level. Do not make up any details about the parameters when you instantiate actions (i.e. if you expect that search results are to be displayed, do not “imagine” an item in the search results and use that as input. Similarly do not make up links, etc.). \\

Here is an example of a bad high-level plan for Task: ``Find the cheapest red t-shirt” \\

High level plan: \\

\begin{Verbatim}[fontsize=\small]
1. (login “example uname”, “example psswd”) 
2. (select “Nike Red T-Shirt 330”)
\end{Verbatim}

Why this is a bad plan: \\ 
- This attempts to log in despite the task not asking for it. \\
- The search result is too specific and is hallucinated.  \\
- If the value is unknown, the action should be abstracted or omitted. \\

A good example of a high-level plan for the same task is: \\
\\
High level plan: 

\begin{Verbatim}[fontsize=\small]
1. (search "red t-shirt")
2. (filter_by_price "low")
3. (select_top_result)
\end{Verbatim}

Below is the task to be accomplished and the web page content: \\
\textbf{Task}: \{task\}   \\
\textbf{Web page content}: \{web\_page\} 

\end{tcolorbox}

\begin{tcolorbox}[width=\columnwidth,
breakable,
title={PDDL Domain Generation},
colback=grey,
colframe=black,
boxrule=0.5pt]

You are given a high level plan in PDDL-like format. Generate a domain file for the high level plan.
You need to generate a domain file for the high level plan. The domain file should be a list of PDDL action templates that describe the proposed plan. \\

\textbf{The task is}: \{task\} \\
\textbf{The high level plan is}: \{ high\_level\_plan \} \\

The domain file should be a list of PDDL action templates that describe the instantiations in the high level plan. \\

Your output should follow the format below exactly. The explanation should come before the answer, and you should use it to explain your reasoning before generating the domain file. \\

\textbf{Explanation}: \\
- Why you chose the preconditions and effects for each action. \\
- Why you chose the parameters for each action. \\
- What each action does. \\

\textbf{Domain File}:

\begin{Verbatim}[fontsize=\small]
(:action login
    :parameters (?u - username ?p - password)
    :precondition (and (on-login-page) 
    (not (logged-in)))
    :effect (and (logged-in))
(:action search_product
    :parameters (?query - string)
    :precondition (and (logged-in))
    :effect (and (search-results-visible))
…
\end{Verbatim}

\textbf{RULES}: You cannot break any of the following rules: \\

- You cannot use markdown formatting. \\
- The domain file should ensure that the high level plan is consistent. The domain file should be written so that no action in the high level plan is called before its preconditions are met. You should make sure that the high level plan is consistent. \\
- Executing an action satisfies its effects. \\
- The initial state of the high level plan is the initial state of the webpage. which is described as "on-homepage" in the domain file. \\
- The action names should be the same as the action names in the high level plan.
\end{tcolorbox}

\subsubsection{Low-level Planning}\label{app:prompt-low-level}

\begin{tcolorbox}[width=\linewidth,
title={low-level planning with action selection},
breakable,
colback=grey,
colframe=black,
boxrule=0.5pt]

You are given a task to perform on a webpage. Propose the next step that is helpful towards achieving the task. \\

You should output only an action that is most likely to help achieve the subgoal. Each action is a dictionary with the following keys: \\

\{
  "INDEX": "action index", 
  "ACTION": "CLICK" | "HOVER" | "TYPE" | "SELECT",
  "SELECTOR": "text='NEWS'" | "button:nth-of-type(3)" | "css=div.menu >> text='Super Bowl'" | ...,
  "VALUE": "if any (You may need to provide a value for the action, e.g. for TYPE action)",
  "TEXT": "the text that is visible on the element",
  "EXPLANATION": "a short explanation of why this action is useful"
\}
\\
Below is the simplified representation of the current state of the webpage: \\
\{web\_page\} \\

History of steps so far: \{history\} \\
Below is the domain file, defining the precondition and postcondition (i.e. the desired outcome) of the high-level goals:\\
\{domain\_file\}\\
Below is the the list of subgoals that have been accomplished so far:\\
\{subgoals\_accomplished\} \\
Below is the next high-level step (subgoal) to be accomplished: \\
Step: \{subgoal\}  \\
Below is the list of available actions. Do not output any actions that are not in the list of available actions. The only field you are allowed to modify is the ``value'' (if a TYPE or SELECT action) and ``explanation'' fields. You are not allowed to modify any other fields. \\

It is very important that your output contains only the action, and nothing else, that is directly parsable as a JSON object.

Available Actions: \\
\{available\_actions\}
\end{tcolorbox}

\begin{tcolorbox}[width=\linewidth,
title={low-level planning with action selection},
breakable,
colback=grey,
colframe=black,
boxrule=0.5pt]
You are an assistant who not only helps to browse and operate web pages to achieve certain goals, but also needs to explore the information on the page to answer the questions raised by the target task. Please answer the following questions as much as possible. \\
        There are key information you will get: \\
        **Key Information**: \\
            - Previous trace: all thoughts, actions and reflections you have made historically. \\
            - Accessibility tree: characteristic expression of the current web page. \\
            
        **Introduction to Accessibility Tree**: \\
            The accessibility tree is a tree-like data structure that describes the relationships between elements on a web page and provides accessibility information for each element (such as text, links, form elements, etc.). \\
            - **Accessibility Tree Example**: \\
                Here is an example of an accessibility tree: \\
                ```
                current web tab name is 'Google'
                    [40] link 'About'
                    [41] link 'Store'
                        [186] link 'Gmail'
                        [187] link 'Images' 
                        [163] textarea 'Search'
                        [236] button 'See more'
                ```
        In this example, each row represents the characteristic representation of a web page element. It has three attributes: '[40]' for the element's element\_id, 'link' indicates the element is a link, and 'About' for the content of the element.
        Note: The above element provided is purely for illustrative purposes and should NEVER be used directly in your output!    \\      

        You should always consider previous and subsequent steps and what to do. \\
        **Thought Space**: \\
            - What action do you think is needed now to complete the task?
            - What's the reason of taking that action?
        
        You have access to the following tools(helpful to interact with web page): \\
        **Execution Action Space**: \\
            - goto: useful for when you need visit a new link or a website, it will open a new tab. \\
            - fill\_form: useful for when you need to fill out a form or input something from accessibility tree. Input should be a string. \\
            - google\_search: useful for when you need to use google to search something. \\
            - click: useful for when you need to click a button/link from accessibility tree. \\
            - select\_option: useful for when you need to select a drop-down box value. When you get (select and option) tags from the accessibility tree, you need to select the serial number(element\_id) corresponding to the select tag, not the option, and select the most likely content corresponding to the option as Input. \\
            - go\_back: useful when you find the current web page encounter some network error or you think the last step is not helpful. \\
            - cache\_data: useful when you need to extract information from the page that you think is extremely valuable for completing the target task. It is not a direct answer to the target task, but it is extremely relevant to the target task. Subsequent actions may refer to this part of the information and return this information as input \\
            - get\_final\_answer: useful for when you think it is the answer to the target task and no other operations are required, Input should be a answer content. \\
        
        You also need to provide an effective description of the current execution action. \\
        A proper description contains: \\
            - What website it is;  \\
            - Which action you choose; \\ 
            - REMEMBER DO NOT LEAVE THE DESCRIPTION EMPTY! \\

        You have to follow the instructions or notes: \\
        **Important Notes**: \\
            - Under the following conditions, you are restricted to using the `google\_search` or `goto` tools exclusively:  \\
                1. In the initial step of a process or when there's no preceding interaction history (i.e., the previous trace is empty). \\
                2. In situations where the accessibility tree is absent or not provided. \\
            - Your action should not be the same as last step's action. \\
            - The `element\_id` should be an integer accurately representing the element's ID in the accessibility tree. \\
            - AVOID using the provided example's element\_id as your output. \\
            - The output JSON blob must be valid; otherwise, it cannot be recognized. \\
        
        **Special Circumstances Guidelines**: \\
            - When performing a search on a website, if you find the search results do not display sufficient content, consider simplifying or modifying your search query. Reducing the complexity of your search query or altering keywords may yield more comprehensive results. \\
        
        Please ensure the accuracy of your output, as we will execute subsequent steps based on the `action`, `action\_input` and `element\_id` you provide. \\
        
        **Output Requirements**: \\
        - Ensure your output strictly adheres to the JSON blob format outlined below:
            
            \{
                "thought": "ACTUAL\_THOUGHT", \\
                "action": "ACTUAL\_TOOLS", \\
                "action\_input": "ACTUAL\_INPUT", \\
                "element\_id": "ACTUAL\_ELEMENT\_ID", \\
                "description": "ACTUAL\_DESCRIPTION"
            \}
          
        - A VALID JSON BLOB EXAMPLE AS FOLLOWS: \\
            \{
                "thought": "In order to complete this task,I need to go to the Google home page", \\
                "action": "click",  \\
                "action\_input": "button", \\
                "element\_id": "236", \\
                "description": "Now I'm on Google's main page. I'm now clicking the button with element\_id [236] to see more information."
            \}

        Current Task Context:
        - Subgoal: \{subgoal\}
        - Previous trace: \{history\}
        - Accessibility tree: \{web\_page\}
 \end{tcolorbox}       

\subsubsection{Postcondition Checking} \label{app:post-check-prompts}

\begin{tcolorbox}[width=\linewidth,
title={Postcondition Checking with Action-ID},
breakable,
colback=grey,
colframe=black,
boxrule=0.5pt]

You are given a high-level step and a web page. Check if the expected effects of the executed action have been achieved. \\
Below is the history of actions performed so far, including the action that was just executed. \\
Each of the actions is a dictionary with the following keys. Think about the effects of each action and what they do to the state of the webpage before deciding. \\
\begin{Verbatim}[fontsize=\small]
"INDEX": "action index", 
"ACTION": "CLICK"|"HOVER"|"TYPE"|"SELECT",
"SELECTOR": "text='NEWS'" 
            | "button:nth-of-type(3)" | ...",
"VALUE": "if any",
"TEXT": "the text visible on the element",
\end{Verbatim}

Each action is also followed by whether it was successful and an observation of the webpage state after the action was executed. \\

Your task: \\
1. Understand and reason about what the executed effects of the high-level step should be. \\
2. Check if the postconditions/effects of the executed action have been achieved based on the history of actions performed so far and the current state of the webpage.  \\
3. Check if the current web page does satisfy the postconditions/effects of the executed action. \\
4. Output an answer to the question: "Have the postconditions/effects of the executed action been achieved?" \\

Your output should ONLY be a json object with the following keys: \\
\{ \\
    "explanation": "a short explanation of why the postconditions were achieved or not, referencing specific elements or changes in the web page" \\
    "answer": "YES" | "NO" \\
\}

Output only The JSON object, and nothing else. Do not use Markdown formatting. 

Below is the task:
\{task\}

Below is the high-level step that we would like to check whether it was successfully achieved:
\{executed\_action\}

[In the case of PDDL] \\
\textbf{Domain File}: \{domain\_file\}

Below is the simplified representation of the current state of the web page:
\{web\_page\} \\

Here is the history of actions performed so far:

\{history\}

\end{tcolorbox}

\begin{tcolorbox}[width=\linewidth,
title={Post-condition checking with expanded action space},
breakable,
colback=grey,
colframe=black,
boxrule=0.5pt]

You are given a high-level step and a web page. Check if the expected effects of the executed action have been achieved. \\

Below is the history of actions performed so far, including the action that was just executed. \\
Each of the actions is a dictionary with the following keys. Think about the effects of each action and what they do to the state of the webpage before deciding. \\
\begin{Verbatim}[fontsize=\small]
"thought": "The reasoning behind taking 
                    this action",
"action": "The action type (goto, click, 
                fill_form, select_option,
    google_search, go_back, cache_data, 
            get_final_answer)",
"action_input": "The input value for 
    the action (e.g. text to fill, 
    URL to navigate to, 
    option to select, etc.)",
"element_id": "The ID of the element 
                to interact with",
"description": "A description of what 
                    the action does"
\end{Verbatim}

Each action is also followed by whether it was successful and an observation of the webpage state after the action was executed. \\

Your task: \\
1. Understand and reason about what the executed effects of the high-level step should be. \\
2. Check if the postconditions/effects of the executed action have been achieved based on the history of actions performed so far and the current state of the webpage.  \\
3. Check if the current web page does satisfy the postconditions/effects of the executed action. \\
4. Output an answer to the question: "Have the postconditions/effects of the executed action been achieved?" \\

Your output should ONLY be a json object with the following keys: \\

\{ \\
    "explanation": "a short explanation of why the postconditions were achieved or not, referencing specific elements or changes in the web page" \\
    "answer": "YES" | "NO" \\
\}

Output only The JSON object, and nothing else. Do not use Markdown formatting. 

Below is the task:
\{task\}

Below is the high-level step that we would like to check whether it was successfully achieved:
\{executed\_action\}

[In the case of PDDL] \\
\textbf{Domain File}: \{domain\_file\}

Below is the simplified representation of the current state of the web page:
\{web\_page\} \\

Here is the history of actions performed so far:

\{history\}

\end{tcolorbox}

\subsubsection{Replanning}
\begin{tcolorbox}[width=\linewidth,
title={Replanning},
breakable,
colback=grey,
colframe=black,
boxrule=0.5pt]

You are an intelligent assistant helping a user complete a task on a web page. \\

IMPORTANT: The user has already attempted this task with a previous plan that did not succeed. You need to propose a NEW high-level plan. \\

You have TWO options: \\
1. **Start over completely**: Propose a completely new high-level plan from scratch, ignoring the previous attempt. \\
2. **Continue from last successful action**: Build upon the actions that were successfully completed, and propose a new plan that continues from where the last successful action left off. \\

\{plan\_context\} \\
\{history\_text\} \\
\{failure\_context\} \\

Based on the history above, decide whether to: \\
- Start over with a completely new approach, OR \\
- Continue from the last successful action \\

Then propose a high-level plan accordingly. \\

First, reflect step-by-step: \\
1. What is the user trying to accomplish? \\
2. What subgoals must be achieved? \\
3. Should I start over or continue from the last successful action? Why? \\
4. Can each subgoal be broken into lower-level actions (click, type, etc.)? \\
5. Are these steps reusable and modular across different pages? \\

Then output your response in this **strict format**. You have to output the plan salvageability assessment and the explanation first, then the proposed high-level plan. \\

Plan Salvageability Assessment: \\
- Is the original plan salvageable? Answer: "YES" or "NO" \\
- Reasoning: Explain why the plan is or is not salvageable. If salvageable, explain which parts can be reused and which steps were already completed. If not salvageable, explain why starting over is necessary. \\
\\
Explanation: \\
- Your reasoning for proposing the high-level plan in 2-3 sentences. \\
- If salvageable, clearly state which steps were already completed and which steps remain. \\
\\
Proposed high-level plan: \\
1. Step one \\
2. Step two \\
...

IMPORTANT REMINDER: If you answered "YES" (plan is salvageable), you MUST ONLY include the REMAINING steps that still need to be completed in your high-level plan. Do NOT repeat steps that were already successfully completed. If you answered "NO" (plan is not salvageable), include ALL steps needed to complete the task from scratch.

Rules: \\
- Do NOT use markdown formatting. \\
- Make sure each step is atomic, reusable, and task-driven. \\
- Focus on clarity, reusability, and structure. \\
- Make sure each step is a high-level skill that can be decomposed into low-level actions and this used to achieve a subgoal needed to achieve the task. \\
- Do not use dummy credentials to log in. Unless explicitly asked to do so as part of the task, and credentials are provided, then you should not propose to sign in. \\
- Do not use dummy inputs. If you do not know the inputs, then you need to find a new action. \\
- You are not allowed to make up specific links. \\
- Make sure each action is atomic and reusable. \\
- Keep the actions at a high level. Do not make up any details about the parameters when you instantiate actions. \\

Webpage: \{web\_page\} \\
Task: \{task\}

\end{tcolorbox}

\subsubsection{End-to-end Evaluation}

\begin{tcolorbox}[width=\linewidth,
title={End-to-end Evaluation},
breakable,
colback=grey,
colframe=black,
boxrule=0.5pt]

You have a task and an action history. Your goal is to evaluate if the action history and the current web page meet the task requirements. You should reason about the action history that got us to the current web page to determine if they meet the task requirements. \\

\{action\_format\_description\} \\

Your answer should be a json object with the following format: \\

\{ \\
    "explanation": "Your explanation here", \\
    "answer": "YES" or "NO", \\
\}

IT IS VERY IMPORTANT THAT YOUR OUTPUT IS A JSON object, AND NOTHING ELSE. THE OUTPUT SHOULD BE DIRECTLY PARSABLE AS A JSON OBJECT. \\

Answer the following question: Does the action sequence meet the task requirements?  \\

**Task**: \{task\} \\
**Action History**: \{action\_history\} \\
**Current Web Page**: \{web\_page\} 

\end{tcolorbox}

\subsubsection{High-level Evaluation}

\begin{tcolorbox}[width=\linewidth,
title={Replanning},
breakable,
colback=grey,
colframe=black,
boxrule=0.5pt]

You are an expert evaluator comparing a human-authored plan and an LLM-generated plan
for the same web navigation task.

TASK: \{task\}

HUMAN PLAN:
\{human\_plan\}

LLM PLAN:
\{llm\_plan\}

For each human step: \\
- Identify which step(s) in the LLM plan correspond to it, if any. Note, that LLM plans might be written in a different syntax or representation of the same step. \\
- Classify its alignment status as one of: \\
  - "aligned": the same intent or action appears in the LLM plan. This means that the step could be matched exactly to one step in the LLM plan. If a step is matched to multiple steps in the LLM plan, then it is decomposed. The step from the LLM plan that is matched to the human step should also not violate or assume additional temporal dependencies that are not present in the human plan. \\
  - "partial": part of the step's intent appears, but it is missing part of the step's intent. \\
  - "missing": the action is completely omitted from the LLM plan. \\
  - "decomposed": the human step is split into multiple smaller LLM steps. This means that the step is matched to multiple steps in the LLM plan. \\
  - "unmatched": If there exists a step in the LLM plan that are not matched to any human step, annotate it as "unmatched". Report all unmatched steps in one entry, with a "null" in the human\_step field and a list of all unmatched steps in the matched\_llm\_steps field. \\

If decomposed: \\
- Determine whether this decomposition is "useful" (adds clarity or logical substructure)
  or "harmful" (introduces redundant or overly detailed low-level steps that assumes preconditions that might not hold true (e.g. if a step requires the existence of a certain button that has not been seen yet)). \\

Return a JSON object of this form: \\
\{ \\
  "steps": [ \\
    \{ \\
        "reasoning": "<brief explanation why the step is aligned, partial, missing, or decomposed and why the steps are matched>",
        "human\_step": "<text>"|null(if reporting unmatched steps),
        "matched\_llm\_steps": ["<matching step(s)>, give the step numbers (1-indexed)"],
        "status": "aligned"|"partial"|"missing"|"decomposed"\\|"unmatched",
    \} \\
  ]
\}

Note that every step in the LLM plan must be matched to a human step or be annotated as "unmatched".

Output JSON only, nothing else.

\end{tcolorbox}

\subsection{Examples of Key Node Annotation}

\begin{figure*}[b]
    \centering
    \begin{subfigure}[t]{\textwidth}
        \centering
        \includegraphics[width=\linewidth]{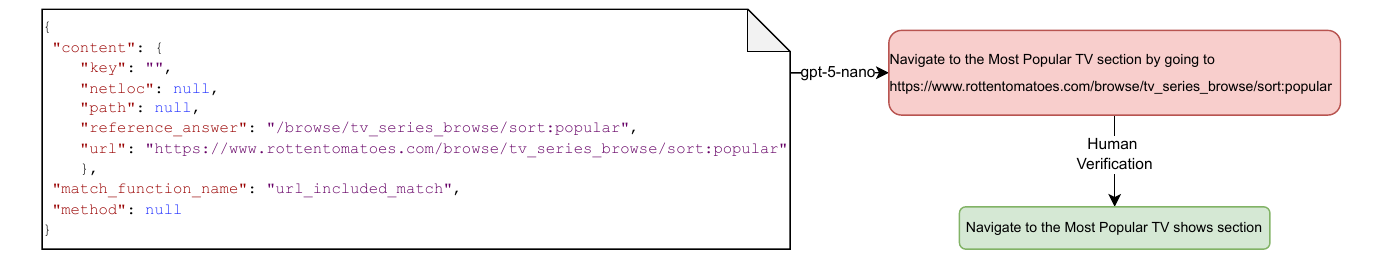}
        \caption{The step proposed by the model is overly specific. Indicating to directly navigate to the specific link.}
        \label{fig:key-node-1}

    \end{subfigure}\hfill
    \begin{subfigure}[t]{\textwidth}
        \centering
        \includegraphics[width=\linewidth]{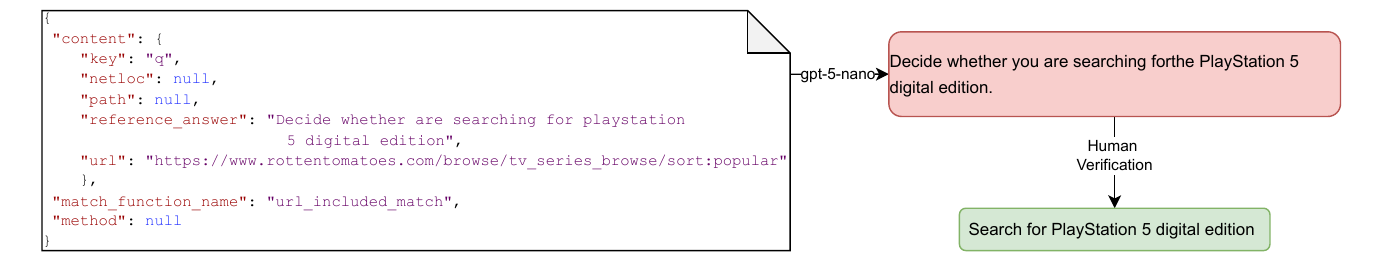}
        \caption{The step proposed by the model is framed as an evaluation, not a step.}
        \label{fig:alignment-tree-b}
    \end{subfigure}
    \caption{\small\textbf{Examples of the high-level step annotation process} We begin by prompting \texttt{gpt-5-nano} to produce a high-level step based on the evaluation key-node object and correct issues such overspecification and steps framed as evaluation functions manually.}
    \label{fig:key-node-examples}
\end{figure*}

The Mind2Web-Live Benchmark \cite{pan2024autonomous} provides a set of manually annotated ``key nodes'' that are used to evaluate task completion; these are assumed by the authors to be necessary subgoals that have to be visited in order for the task to be completed. We prompt an LLM to produce high-level steps based on these evaluation objects. Examples of the outputs and the issues we encounter are presented in Figure \ref{fig:key-node-examples}. 

\subsection{The Reliability of LLM-as-Judge}\label{app:judge-reliability}

LLM-as-Judge evaluation is becoming increasingly popular across many agentic applications \cite{xue2025illusionprogressassessingcurrent, gou2025mindweb2}. This is because of the convenience of LLMs' ability to leverage its language understanding and the  knowledge acquired during training to understand, interpret and reason about open-ended requirements and unstructured data. This was the motivation for using LLM-as-Judge for both postcondition-checking and final evaluation. To establish the reliability of these methods, we sample 50 examples of postcondition checks and final evaluations and perform manual annotation, then compare the outcomes. We find that we agree with the postcondition checks in 41 out 50 instances (82\%) and with the final evaluation in 43 out of 50 instances (86\%), indicating that using LLM-as-Judge can be effective as it exhibits a high agreement rate with human evaluation. 

\subsection{High-level Alignment Evaluation Trees}\label{app:alignment-trees}
\begin{figure}[H]
    \centering
    \begin{subfigure}[t]{\linewidth}
        \centering
        \includegraphics[width=\linewidth]{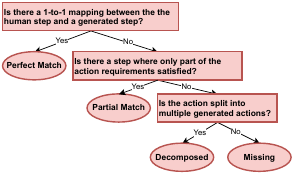}
        \caption{Are the human steps captured by the LLM-generated plan?}
        \label{fig:alignment-tree-a}

    \end{subfigure}\hfill
    \begin{subfigure}[t]{\linewidth}
        \centering
        \includegraphics[width=\linewidth]{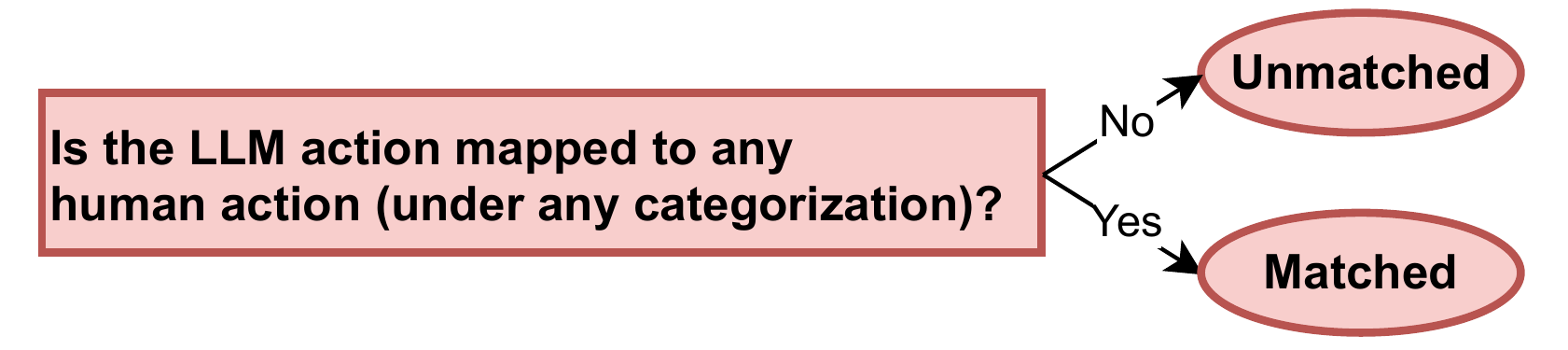}
        \caption{Are the LLM-generated steps relevant to the human plan?}
        \label{fig:alignment-tree-b}
    \end{subfigure}
    \caption{{Evaluation trees for high-level alignment.}}
    \label{fig:alignment-tree}
\end{figure}

Figure \ref{fig:alignment-tree} presents two decision trees used to evaluate alignment between human-authored plans and LLM-generated plans at the high level.

The first tree shown in Figure \ref{fig:alignment-tree-a} examines whether each human-authored step is captured by the LLM-generated plan. It first checks for a one-to-one correspondence between a human step and a generated step, which is labeled as a Perfect Match. If no such correspondence exists, the tree determines whether the generated step only partially satisfies the human step, resulting in a Partial Match. If not partial, it then checks whether the human step is split into multiple finer-grained generated steps, labeled as Decomposed. If none of these conditions are met, the human step is classified as Missing, indicating it is absent from the LLM-generated plan. The second tree in Figure \ref{fig:alignment-tree-b} evaluates the relevance of the LLM-generated steps to the human plan. Each generated step is checked to see whether it maps to any human-authored step under any of the categories from the first tree. Generated steps that align with a human step are labeled Matched, while those that do not correspond to any human step are labeled Unmatched, reflecting redundant or irrelevant actions.

\subsection{Example of NL and PDDL plans} \label{app:nlvspddl}
In Figure~\ref{fig:alignment-tree}, we present motivating example highlighting the importance of high-level representation. NL plans can be overly specific and makes assumptions about the environment that may not hold true. Some of the steps can also not easily be translated into a sequence of actions. PDDL alleviates this by providing better abstraction of the same high-level plan.

\begin{figure}[H]
    \centering
    \includegraphics[width=\linewidth]{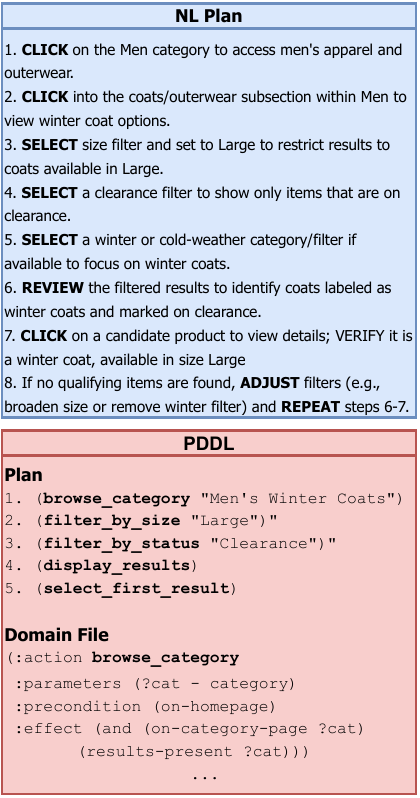}
    \caption{\textbf{Motivating example highlighting the importance of high-level representation.} }
    \label{fig:nlvspddl}
    \vspace{-0.15in}
\end{figure}

\section{Additional Analysis}\label{app:aditional-analysis}

\subsection{Full Hierarchical Planning Example}

\begin{figure*}[t]
\begin{tcolorbox}[colback=gray!3, colframe=black!40,
  title=Example hierarchical web planning instance]
\footnotesize

\textbf{Task.} \emph{``Find me the deals available for the Great escape park on sixflags''}

\vspace{4pt}
\centering
\includegraphics[width=0.7\linewidth]{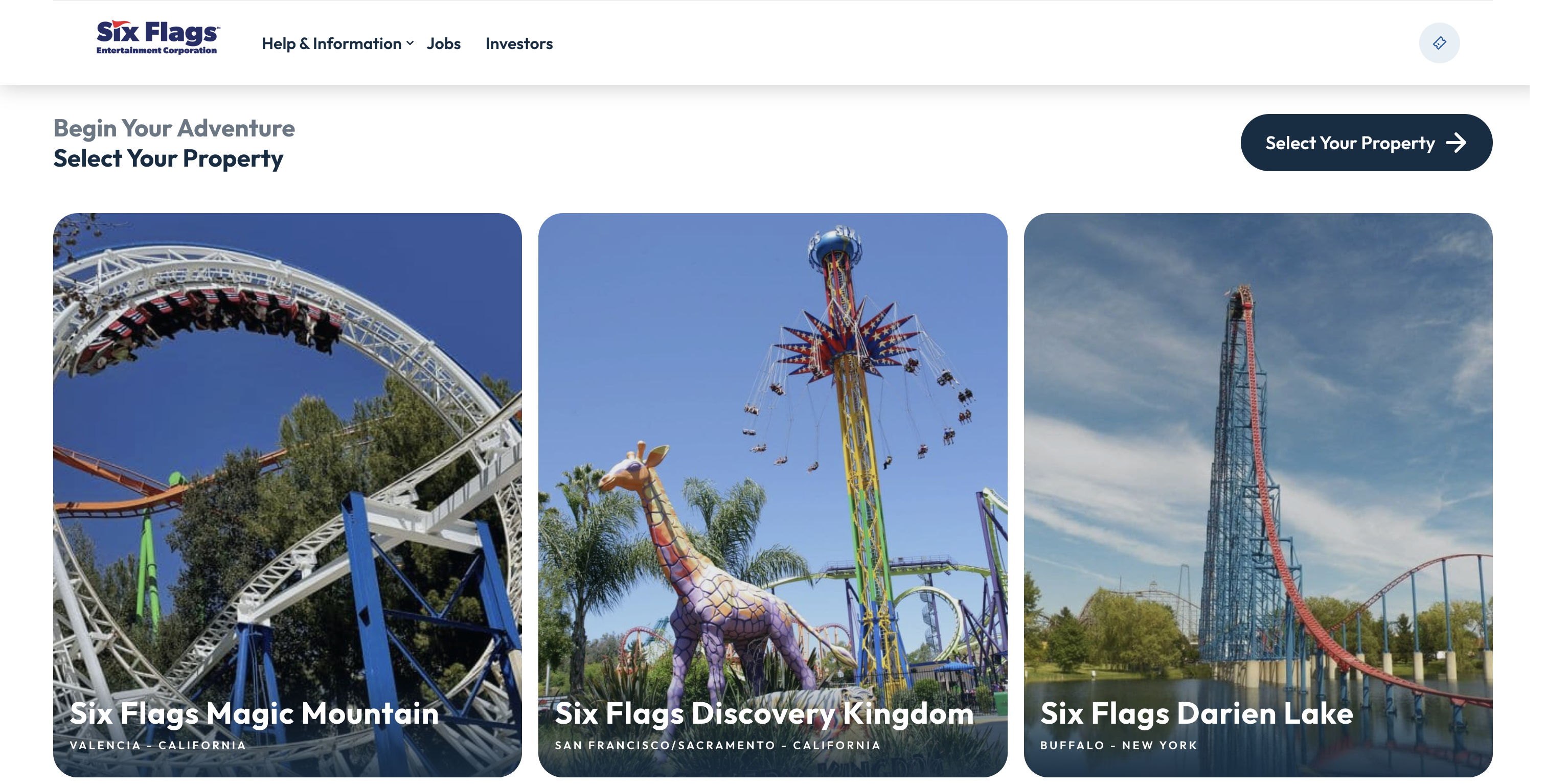}\\
\caption*{\small Initial page shown to the agent (homepage of Six Flags).}

\vspace{2pt}
\hrule
\vspace{2pt}

\noindent\textbf{High-level planning}\\[1pt]
\begin{minipage}[t]{0.32\linewidth}
\textbf{Human plan}\\[2pt]
1. Navigate to the Great Escape park page.\\
2. Go to the ticket store section for the Great Escape park.
\end{minipage}
\hfill
\begin{minipage}[t]{0.32\linewidth}
\textbf{LLM NL plan}\\[2pt]
1. SELECT the Great Escape park as the active park using the Park of Preference control.\\
2. Locate the Deals or Promotions section for the Great Escape park and CLICK it if present. \\
3. If a Deals/Promotions section exists, READ and RECORD all current deals offered for Great Escape.\\
4. If no Deals/Promotions section is found, CLICK into Tickets or Prices sections and SCAN for any current promotions or bundled offers, then RECORD any found promos.\\
5. Compile and present a concise report listing all deals, including name, discount, eligibility, and how to redeem, and note if no deals are available with suggested next steps.

\end{minipage}
\hfill
\begin{minipage}[t]{0.32\linewidth}
\textbf{LLM PDDL plan}\\[2pt]

1. (open\_deals\_section) \\
2. (select\_park "Great Escape Park")\\
3. (extract\_deals)

\end{minipage}

\vspace{2pt}
\hrule
\vspace{2pt}

\noindent\textbf{Low-level planning (LLM) [Excerpt for subgoal: (select\_park "Great Escape Park")]}\\[1pt]
\begin{minipage}[t]{\linewidth}
\textbf{Action:} google\_search "Six Flags deals site:sixflags.com"  
\textbf{Observation:} SUCCESS — Google search executed.

\textbf{Action:} click "Deals \& Offers - Six Flags Great Adventure"  
\textbf{Observation:} SUCCESS — link clicked.

\textbf{Action:} goto "https://www.sixflags.com/national-sites"  
\textbf{Observation:} SUCCESS — navigated to National Sites page.

\textbf{Action:} select\_option "Great Escape Park"  
\textbf{Observation:} FAILED — could not select option.

\textbf{Action:} click "Select a Park" combobox  
\textbf{Observation:} SUCCESS — combobox clicked.

\textbf{Action:} select\_option "Great Escape Park"  
\textbf{Observation:} FAILED — could not select option.

\textbf{Action:} click "Select a Park" combobox  
\textbf{Observation:} SUCCESS — combobox clicked.

\textbf{Action:} select\_option "Great Escape Park"  
\textbf{Observation:} FAILED — could not select option.

\end{minipage}

\vspace{2pt}
\hrule 
\vspace{2pt}

\noindent\textbf{Postcondition checking for subgoal (select\_park ``Great Escape Park'')}\\[1pt]
\begin{minipage}{\linewidth}
\textbf{Postconditions not met for action: 2. (select\_park "Great Escape Park")}. The postconditions of select\_park('Great Escape Park') were not achieved. The park was never successfully selected from the park selector. Multiple attempts to select 'Great Escape Park' failed with 'Could not select 'Great Escape Park'', so neither park-selected Great Escape Park nor deals-loaded-for Great Escape Park were achieved.
\end{minipage}

\vspace{2pt}

\vspace{2pt}

\noindent\textbf{Final Evaluation (PDDL trace)}\\[1pt]
\begin{minipage}{\linewidth}
\textbf{Answer:} No\\
\textbf{Explanation:} The action history shows that the user has attempted to find the deals for Great Escape Park on the Six Flags website by performing several actions, including navigating to the National Sites page and trying to select Great Escape Park from the park selector. However, the actions to select the park have failed multiple times, and the current web page is still the homepage of Six Flags, not the deals page for Great Escape Park. Therefore, the task requirements are not met as no deals have been found yet.
\end{minipage}
\end{tcolorbox}
\caption{Example of our hierarchical evaluation pipeline for a single task, showing high-level planning (Human vs.\ NL vs.\ PDDL), low-level execution behavior, postcondition checking, and the resulting evaluation signal (\texttt{gpt-5-nano}).}
\label{fig:example-hierarchical}
\vspace{-0.1in}
\end{figure*}

Figure \ref{fig:example-hierarchical} shows a running examples of the hierarchical planning framework we evaluate in this work. 

\subsection{Distribution of low-level Actions Selected}\label{app:action-distribution}

\begin{figure}[H]
    \centering
    \begin{subfigure}{\linewidth}
        \centering
        \includegraphics[width=\linewidth]{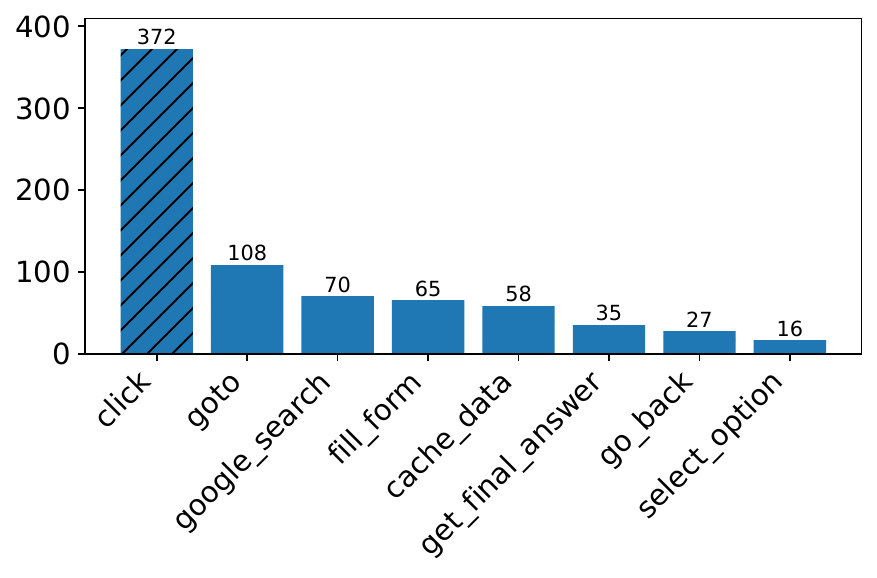}
        \caption{\texttt{gpt-5-nano}}
    \end{subfigure}

    \begin{subfigure}{\linewidth}
        \centering
        \includegraphics[width=\linewidth]{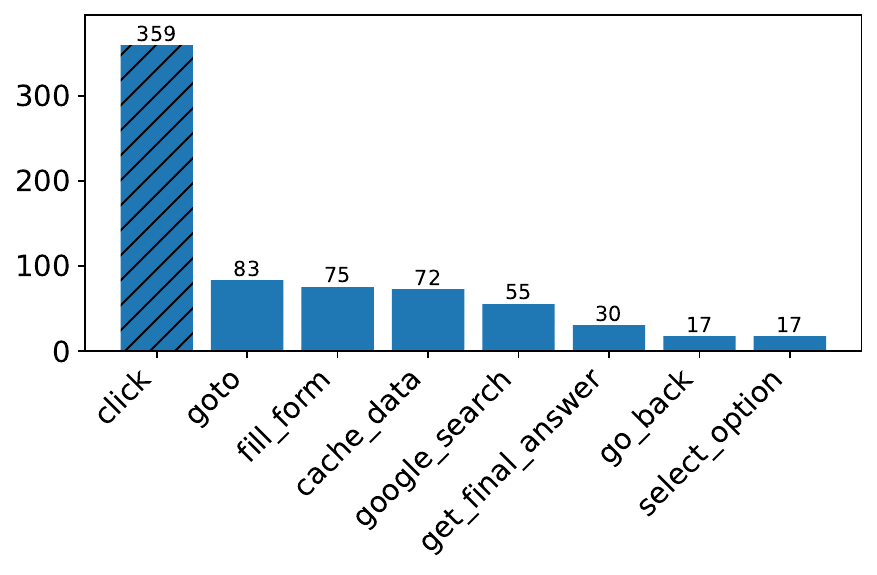}
        \caption{\texttt{claude-haiku-4.5}}
    \end{subfigure}

    \begin{subfigure}{\linewidth}
        \centering
        \includegraphics[width=\linewidth]{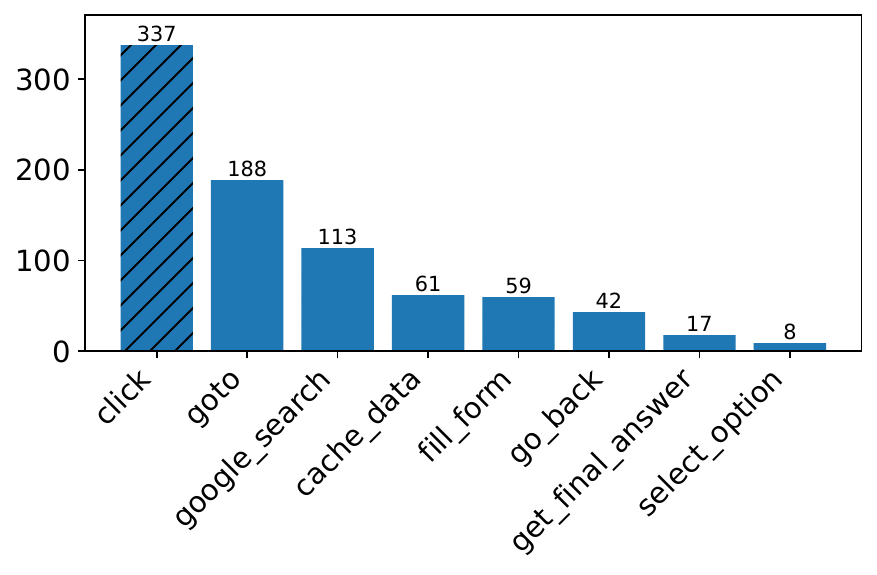}
        \caption{\texttt{gemini-flash-4.5}}
    \end{subfigure}

    \caption{Distribution of actions in the expanded action space across different LLMs.}
    \label{fig:action-distribution}
\end{figure}

The frequency of each of the actions from the expanded being selected by each model is shown in Figure \ref{fig:action-distribution}.

\textbf{LLM-based executors perfer clicks and direct navigation}, as shown in Figure ~\ref{fig:action-distribution}, they heavily favor \textit{click}, \textit{goto}, and \textit{google\_search} actions. While this can help escape local dead ends, it often leads to off-task behavior; direct links can lead to non-existent pages, and \textit{google\_search} actions frequently redirect the agent outside the target site (e.g., retrieving information from Wikipedia instead of ESPN as shown in Figure \ref{fig:error-analysis}). Table ~\ref{tab:failure-modes} shows that over 16\% of actions are performed outside the required website domain, and 32\% of \texttt{goto} actions lead to non-existing links. 

\subsection{Examples of low-level failures}\label{app:low-level-failures}

\begin{figure*}[tb]
    \centering
    \includegraphics[width=\textwidth]{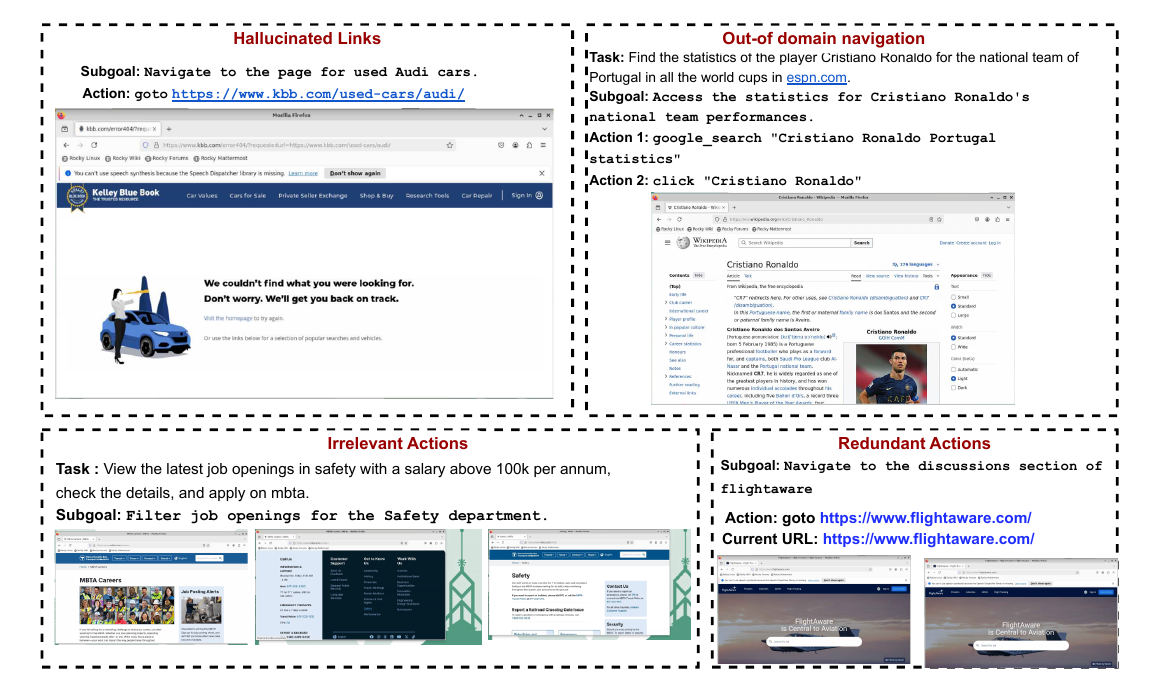}
    \caption{Examples of some of the most common errors encountered by the model (\texttt{gpt-5-nano})}
    \label{fig:error-analysis}
\end{figure*}

Figure \ref{fig:error-analysis} shows examples of some of the most common errors encountered by the model. 

\noindent\emph{Low-level executors can misinterpret subgoals or confuse visually similar elements,} for example, filtering for the ``Safety department'' but selecting a button labeled ``Safety'' instead, which leads to a page on the safety practices of the company. Models also make incorrect assumptions about task completion, such as entering a ZIP code without validating the search result. 

\subsection{Performance of Other Models} \label{app:other-models}

\subsubsection{\texttt{claude-haiku-4.5}}
\begin{figure}[H]
    \centering
    \includegraphics[width=\linewidth]{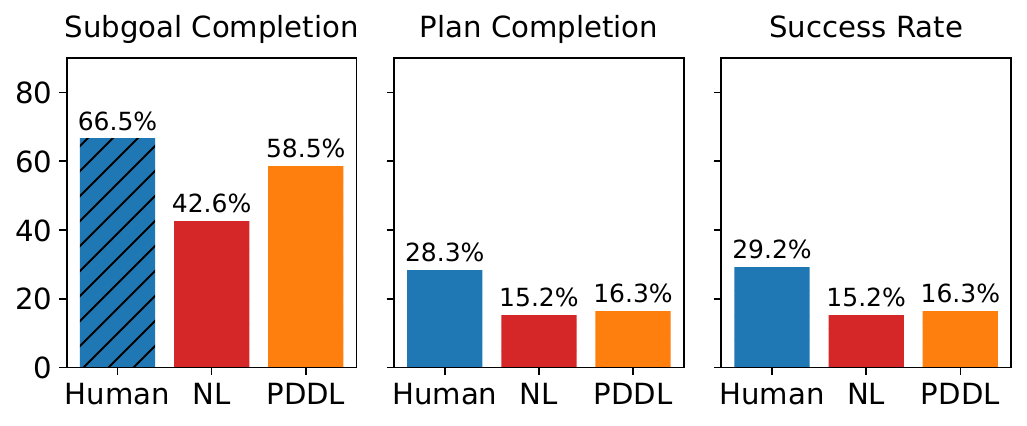}
    \caption{Execution Results using different high-level representations using \texttt{claude-haiku-4.5}}
    \label{fig:execution-claude}
\end{figure}

\begin{figure}[H]
    \centering
    \includegraphics[width=\linewidth]{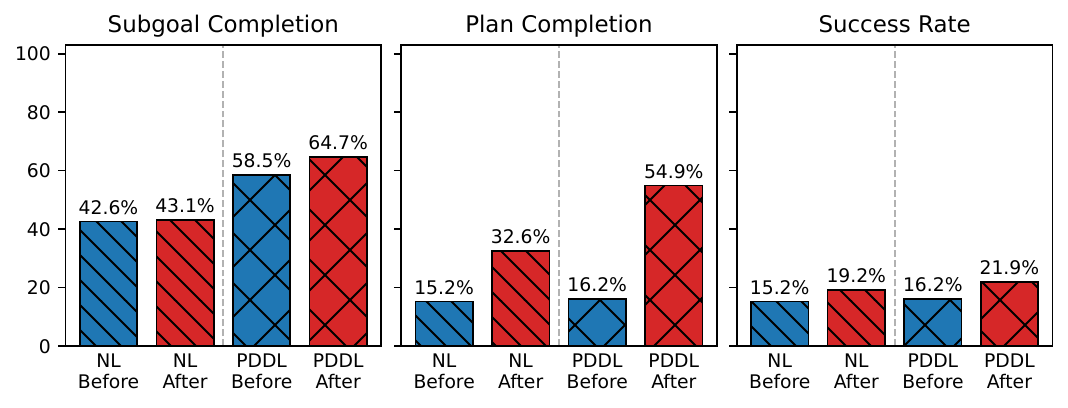}
    \caption{Replanning results \texttt{claude-haiku-4.5}}
    \label{fig:claude-replanning}
\end{figure}

The performance of \texttt{claude-haiku-4.5} across each of the three levels is illustrated below. Table~\ref{tab:claude-high-level} illustrates the peformance on high-level alignment metrics. Table~\ref{tab:efficiency-claude} summarizes the efficiency of the model in terms of number of actions and subgoals. Figure \ref{fig:execution-claude} showcases execution results for this model. Failure mode statistics are summarized in Table~\ref{tab:failure-modes-claude}, while replanning results are summarized in Figure \ref{fig:claude-replanning}.

\begin{table}[H]
    \centering
    
    \small
    \begin{tabular}{lcc}
        \toprule
        & \textbf{NL} & \textbf{PDDL} \\
        \midrule
        \textbf{Perfect Match} & 54.9 & 66.7  \\
        \textbf{Partial}       & 6.8 & 4.8   \\
        \textbf{Missing}       & 7.5 & 7.3 \\
        \textbf{Decomposed}    & 30.8 & 21.2 \\
        \midrule
        \textbf{Unmatched}     & 17.4 & 10.1 \\
        \textbf{Matched}       & 83.6 & 89.9 \\
        \bottomrule
    \end{tabular}%
    \caption{Alignment (\%) between human and LLM-generated high-level plans: \texttt{claude-haiku-4.5}.}\label{tab:claude-high-level}
\end{table}

\begin{table}[H]
    \centering
    \small
    \begin{tabular}{lcc}
        \toprule
         & \textbf{NL} & \textbf{PDDL} \\
        \midrule
        \textbf{\# High-level steps} & 4.89 & 3.70  \\
        \textbf{\# Low-level actions} & 11.0 & 10.6 \\
        \bottomrule
    \end{tabular}
    \caption{Avg. Length of High-level and Low-level plans for each representation for \texttt{claude-haiku-4.5}}
    \label{tab:efficiency-claude}
    \vspace{-0.15in}
\end{table}

\begin{table}[H]
    \centering
    \small
        \begin{tabular}{lcl}
            \toprule
            \textbf{Failure Mode} & \textbf{Rate (\%)} & \textbf{Base Category} \\
            \midrule
            Repetitions & 16.7 & of failures \\
            Hallucinated links & 25.3 & of \texttt{goto} actions \\
            Redundant & 23.6 & of all actions \\
            Out-of-domain links & 16.9 & of all states \\
            \bottomrule
        \end{tabular}
    \caption{\textbf{Failure modes across different categories.}
    Each rate is computed relative to its own base set. \texttt{claude-haiku-4.5}}
    \label{tab:failure-modes-claude}

\end{table}

\subsubsection{\texttt{gemini-flash-2.5}}

\begin{figure}[H]
    \centering
    \includegraphics[width=\linewidth]{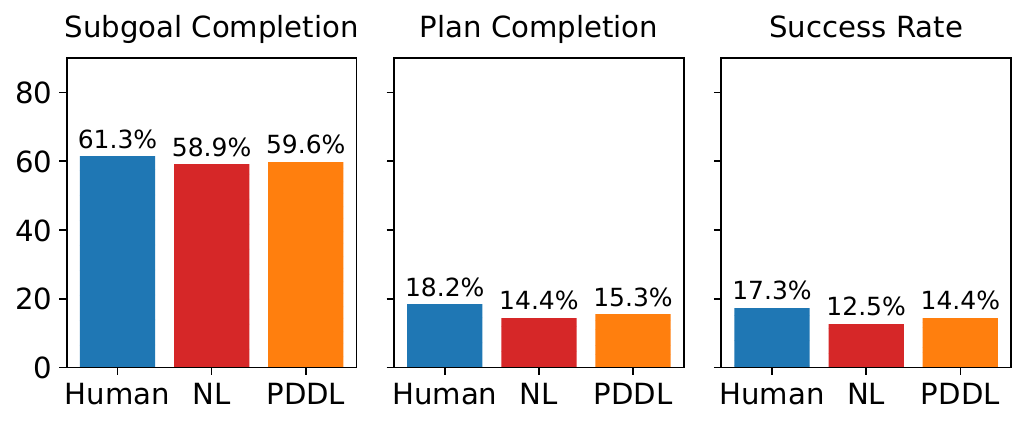}
    \caption{Execution Results using different high-level representations using \texttt{gemini-flash-2.5}}
    \label{fig:execution-gemini}
\end{figure}

\begin{figure}[H]
    \centering
    \includegraphics[width=\linewidth]{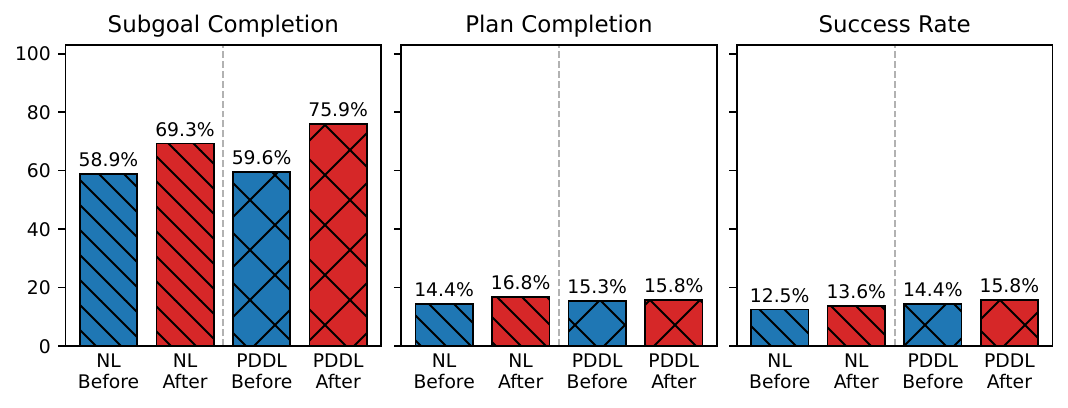}
    \caption{Replanning results \texttt{gemini-flash-2.5}}
    \label{fig:gemini-replanning}
\end{figure}

The performance of \texttt{gemini-flash-2.5} across each of the three levels is illustrated below. Table~\ref{tab:gemini-high-level} illustrates the peformance on high-level alignment metrics. Table~\ref{tab:efficiency-claude} summarizes the efficiency of the model in terms of number of actions and subgoals. Figure \ref{fig:execution-gemini} showcases execution results for this model. Failure mode statistics are summarized in Table~\ref{tab:failure-modes-gemini}, while replanning results are summarized in Figure \ref{fig:gemini-replanning}.

\begin{table}[H]
    \centering
    \small
    \begin{tabular}{lcc}
        \toprule
        & \textbf{NL} & \textbf{PDDL} \\
        \midrule
        \textbf{Perfect Match} & 61.4 & 64.3 \\
        \textbf{Partial}       & 8.5  & 7.2  \\
        \textbf{Missing}       & 11.6 & 12.4 \\
        \textbf{Decomposed}    & 18.5 & 16.1 \\
        \midrule
        \textbf{Unmatched}     & 21.6 & 18.3 \\
        \textbf{Matched}       & 78.4 & 81.7 \\
        \bottomrule
    \end{tabular}%
    \caption{Alignment (\%) between human and LLM-generated high-level plans: \texttt{gemini-flash-2.5}.}\label{tab:gemini-high-level}
\end{table}

\begin{table}[H]
    \centering
    \resizebox{\columnwidth}{!}{
        \begin{tabular}{lcl}
            \toprule
            \textbf{Failure Mode} & \textbf{Rate (\%)} & \textbf{Base Category} \\
            \midrule
            Repetitions & 6.70 & of failures \\
            Hallucinated links & 31.2 & of \texttt{goto} actions \\
            Redundant & 41.2 & of all actions \\
            Out-of-domain links & 14.7 & of all actions \\
            \bottomrule
        \end{tabular}
    }
    \caption{\textbf{Failure modes across different categories.}
    Each rate is computed relative to its own base set. \texttt{gemini-flash-2.5}}
    \label{tab:failure-modes-gemini}
\end{table}

\begin{table}[H]
    \centering
    \small
    \begin{tabular}{lcc}
        \toprule
        & \textbf{NL} & \textbf{PDDL} \\
        \midrule
        \textbf{\# High-level steps} & 3.63 & 3.19  \\
        \textbf{\# Low-level actions} & 8.55 & 8.25 \\
        \bottomrule
    \end{tabular}
    \caption{Avg. Length of High-level and Low-level plans for each representation for \texttt{gemini-flash-2.5}}
    \label{tab:efficiency-gemini}
    \vspace{-0.15in}
\end{table}

\end{document}